\documentclass{article}

\usepackage[final]{neurips_2022}

\usepackage[utf8]{inputenc} 
\usepackage[T1]{fontenc}    
\usepackage{hyperref}       
\usepackage{url}            
\usepackage{booktabs}       
\usepackage{amsfonts}       
\usepackage{nicefrac}       
\usepackage{microtype}      
\usepackage{xcolor}         

\usepackage{graphicx}
\usepackage{amsmath}
\usepackage{amssymb}
\usepackage{hhline}
\usepackage{multirow}
\usepackage[ruled,linesnumbered]{algorithm2e}
\newcommand{\E}{\mathbb{E}}
\hypersetup{
    colorlinks=true,
    linkcolor=blue,
    filecolor=blue,      
    urlcolor=blue,
    citecolor=blue,
}
\usepackage{subcaption}

\title{Confidence-based Reliable Learning \\under Dual Noises}

\author{%
  Peng Cui\textsuperscript{1 3}, Yang Yue\textsuperscript{1}, Zhijie Deng\textsuperscript{1 2}\thanks{The corresponding author.}\,\,, Jun Zhu\textsuperscript{1}\footnotemark[1] \\
\textsuperscript{1} Dept. of Comp. Sci. \& Tech., Institute for AI, BNRist Center, \\
Tsinghua-Bosch Joint ML Center, THBI Lab, Tsinghua University, Beijing, 100084 China \\ 
\textsuperscript{2} Qing Yuan Research Institute, Shanghai Jiao Tong University \quad \textsuperscript{3} RealAI\\
 \texttt{xpeng.cui@gmail.com,~yueyang22@mails.tsinghua.edu.cn}\\
 \texttt{zhijied@sjtu.edu.cn,~dcszj@tsinghua.edu.cn}
}

\begin{document}

\maketitle

\begin{abstract}
  Deep neural networks (DNNs) have achieved remarkable success in a variety of computer vision tasks, where massive labeled images are routinely required for model optimization. Yet, the data collected from the open world are unavoidably polluted by noise, which may significantly undermine the efficacy of the learned models. Various attempts have been made to reliably train DNNs under data noise, but they separately account for either the noise existing in the labels or that existing in the images. A naive combination of the two lines of works would suffer from the limitations in both sides, and miss the opportunities to handle the two kinds of noise in parallel. This work provides a first, unified framework for reliable learning under the joint (image, label)-noise. Technically, we develop a confidence-based sample filter to progressively filter out noisy data without the need of pre-specifying noise ratio. Then, we penalize the model uncertainty of the detected noisy data instead of letting the model continue over-fitting the misleading information in them. Experimental results on various challenging synthetic and real-world noisy datasets verify that the proposed method can outperform competing baselines in the aspect of classification performance. 
\end{abstract}

\section{Introduction}
\label{sec:intro}
Deep Neural Networks (DNNs) have obtained great success in a wide spectrum of computer vision applications~\cite{deep2015,ren2015faster,he2016deep,he2017mask}, especially when a large volume of carefully-annotated low-distortion images are available. 
However, the images collected from the wild in real-world tasks are unavoidably polluted by noise in the images themselves (e.g., image corruptions~\cite{hendrycks2018benchmarking} and background noise~\cite{tu2020learning}) or the associated labels~\cite{natarajan2013learning}, termed as image noise (\emph{x-noise}) and label noise (\emph{y-noise}) respectively. 
Previous investigations show that the DNNs naively trained under \emph{y-noise}~\cite{arpit2017closer,zhang2021understanding} or \emph{x-noise}~\cite{dodge2016understanding,zhou2017classification} suffer from detrimental over-fitting issues, thus exhibit poor generalization performance and serious over-confidence.
There has been a large body of attempts towards dealing with data noise, but they mainly focus on a limited setting, where noise only exists in either the label (i.e., noisy labels)~\cite{natarajan2013learning,arazo2019unsupervised,liu2020peer,cheng2020learning} or the image~\cite{fergus2006removing,levin2009understanding,xu2014deep}.
It is non-trivial to extend them to exhaustively deal with dual noises (i.e., the joint \emph{(x,y)-noise}). 
Moreover, the techniques for handling \emph{x-noise} suffer from non-trivial limitations. For example, most image denoising methods work on well-preserved image texture~\cite{fan2019brief}, thus may easily fail when facing images that are globally blurred (see Fig.~\ref{deblur} in Appendix); alternative image Super-Resolution (SR) solutions are usually computationally expensive~\cite{wang2020deep}. 
These issues raise the requirement of a unified approach for reliable learning under dual noises.

Compared to deterministic DNNs, uncertainty-based deep models (e.g., Bayesian Neural Networks (BNNs)~\cite{blundell2015weight} and \emph{deep ensemble}~\cite{Lakshminarayanan}) reason about the uncertainty and hence have the potential to mitigate the over-fitting to noisy data. 
Empowered by this insight, we first perform a systematical investigation on leveraging uncertainty-based deep models to cope with dual noises. 
We observe that, despite with less over-fitting, the uncertainty-based deep models may still suffer from the bias in the noisy data and yield compromising results.

To further ameliorate the pathologies induced by data noise and achieve reliable learning, we propose a novel workflow for the learning of uncertainty-based deep models under dual noises. 
Firstly, inspired by the recent success of using predictive confidence to detect the out-of-distribution data~\cite{hendrycks2016baseline}, we propose to detect both the noisy images and the noisy labels by the predictive confidence produced by uncertainty-based deep models. 
Concretely, we use the predictive probability corresponding to the label (i.e., label confidence) to filter out the samples with \emph{y-noise}, and use the maximum confidence to filter out the samples with \emph{x-noise}.
After doing so, we propose to penalize the uncertainty~\cite{Kendall} of the detected noisy data to make use of the valuable information inside the images without relying on the misleading supervisory information.

Given the merits of \emph{deep ensemble}~\cite{Lakshminarayanan} for providing calibrated confidence and uncertainty under distribution shift revealed by related works ~\cite{ovadia2019can} and our studies, we opt to place our workflow on deep ensemble to establish a strong, scalable, and easy-to-implement baseline for learning under dual noises. 
Of note that the developed strategies are readily applicable to other uncertainty-based deep models like BNNs.

We perform extensive empirical studies to evidence the effectiveness of the proposed method. 
We first show that the proposed method significantly outperforms competitive baselines on CIFAR-100 and TinyImageNet datasets with different levels of synthetic \emph{(x,y)-noise}. 
We then verify the superiority of the proposed method on the challenging WebVision benchmark~\cite{li2017webvision} which contains extensive real-world noise.  
We further provide insightful ablation studies to show the robustness of our approach to multiple hyper-parameters.
\section{Related Work}
Many methods have been proposed to deal with \emph{y-noise} in deep learning.
A direct approach is to design the robust loss functions, e.g., the loss function based on the mean absolute error~\cite{ghosh2017robust} and the symmetric cross-entropy~\cite{wang2019symmetric,charoenphakdee2019symmetric}. However, it is challenging to deal with the noisy data with high noise rates. An alternative method is to train on reweighing or selected training examples, e.g., estimating the weight of samples based on meta-learning~\cite{han2018co}, MentorNet~\cite{jiang2018mentornet} and Co-teaching~\cite{ren2018learning}, but designing an effective algorithm or criterion of selecting the samples based on the deterministic DNNs tends to be difficult. Recently, the loss correction approaches are also used to mitigate the over-fitting to noisy labels by assigning a weight to the prediction of the model~\cite{reed2014training,arazo2019unsupervised} or by adding a regularization to the loss function~\cite{liu2020peer,cheng2020learning}. 
To deal with \emph{x-noise}, image denoising may be a useful technique. \cite{fergus2006removing} assumes a uniform camera blur over the image and then applies a standard deconvolution algorithm to reconstruct the blurry image, but it can only handle those locally-blurred images.  ~\cite{xu2014deep,mao2016image} propose to use a deep convolutional neural network to capture the characteristics of degradation and restorate blurred images, but they commonly need image pairs (i.e., the label indicates clean or noisy) for training and the supervised information cannot be provided in our setting.
Therefore, it is not free to extend the existing works to handle dual noises, and developing new techniques is necessary.

Typically, in machine learning and computer vision, the uncertainty we are concerned about can be classified into two categories: \emph{Epistemic} uncertainty and \emph{Aleatoric} uncertainty, which are also called model uncertainty and data uncertainty~\cite{Kendall}. Extensive uncertainty quantification approaches have been proposed in the literature. A direct approach to incorporating uncertainty into DNNs is to perform Bayesian inference over DNN weights, with BNNs~\cite{blundell2015weight,liu2016stein}, Monte Carlo (MC) dropout~\cite{gal2016dropout} and SWAG~\cite{maddox2019simple} as popular examples.
Yet, Bayesian inference is often expensive due to the high non-linearity of DNNs. An alternative way is to adapt various distance-aware output layers into DNNs in a non-Bayesian way~\cite{van2020uncertainty,liu2020simple,malinin2018predictive}. However, these methods may suffer from degenerated uncertainty estimates~\cite{fort2019deep} due to restrictive assumptions. \emph{Deep ensemble}~\cite{Lakshminarayanan} is a prevalent and leading tool for uncertainty quantification, which can produce calibrated confidence and uncertainty~\cite{ovadia2019can} by assembling the outputs of different DNN predictors uniformly. In this paper, we place our workflow on deep ensemble to establish a robust learning approach under dual noises.
\begin{figure}
  \centering
  \begin{subfigure}[1]{0.32\linewidth}
    \includegraphics[width=0.95\linewidth]{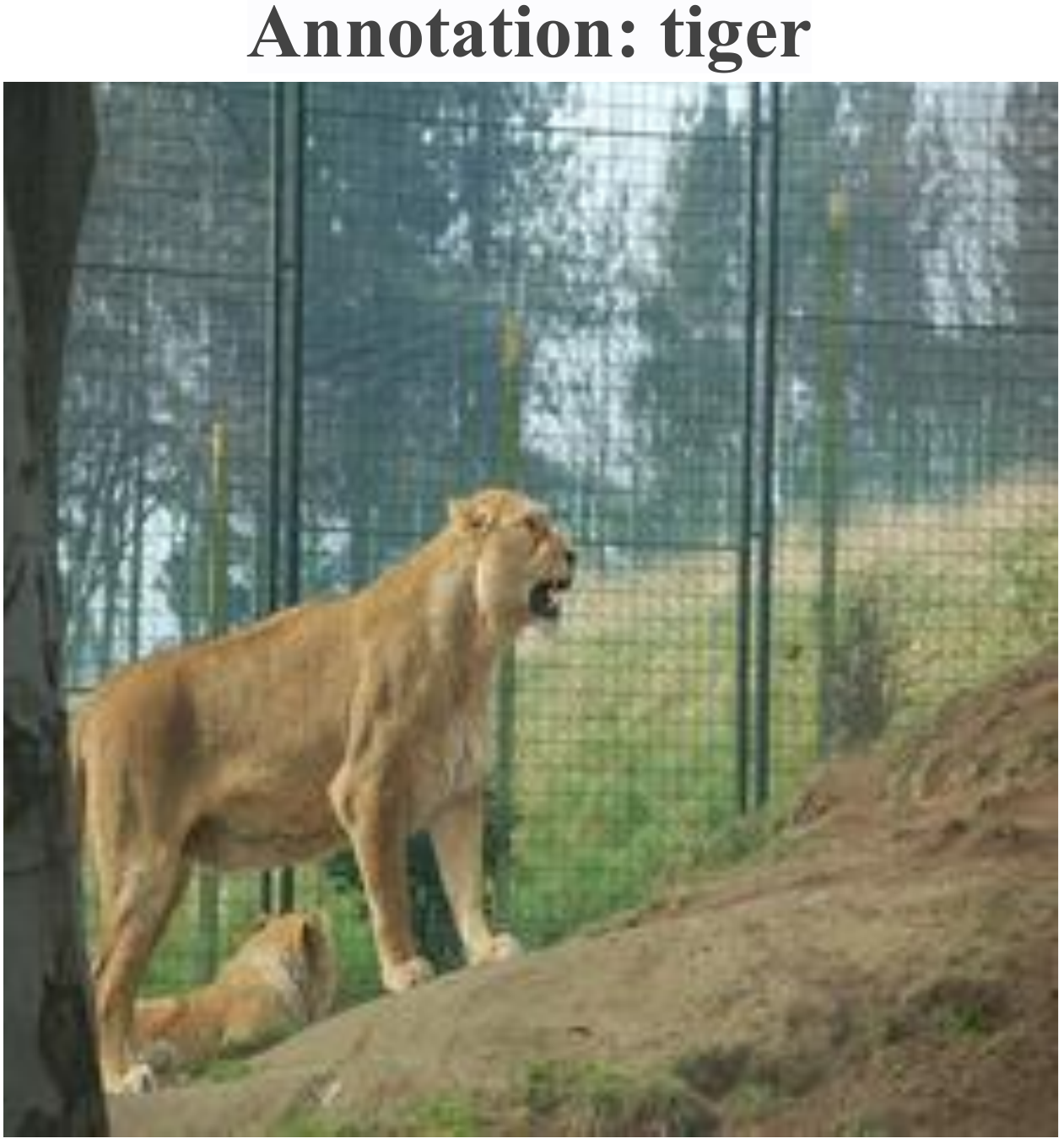}
    \caption{\emph{y-noise}: wrong annotations.}
    \label{fig:lnoise}
  \end{subfigure}%
  \hfill
  \begin{subfigure}[2]{0.32\linewidth}
    \includegraphics[width=0.95\linewidth]{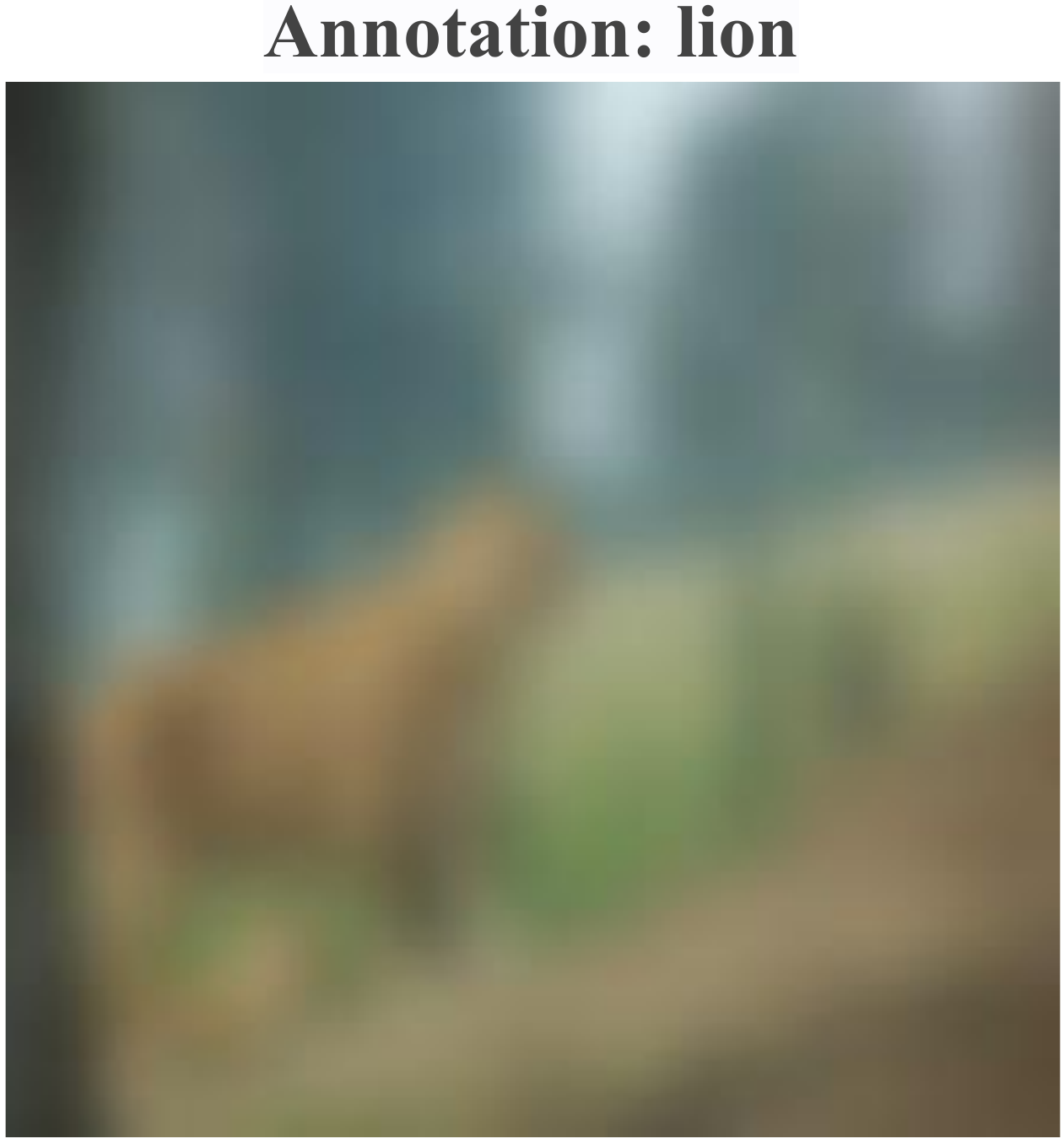}
    \caption{\emph{x-noise} \uppercase\expandafter{\romannumeral1}: corrupted images.}
    \label{fig:xnoise1}
  \end{subfigure}
  \begin{subfigure}[3]{0.32\linewidth}
    \includegraphics[width=0.95\linewidth]{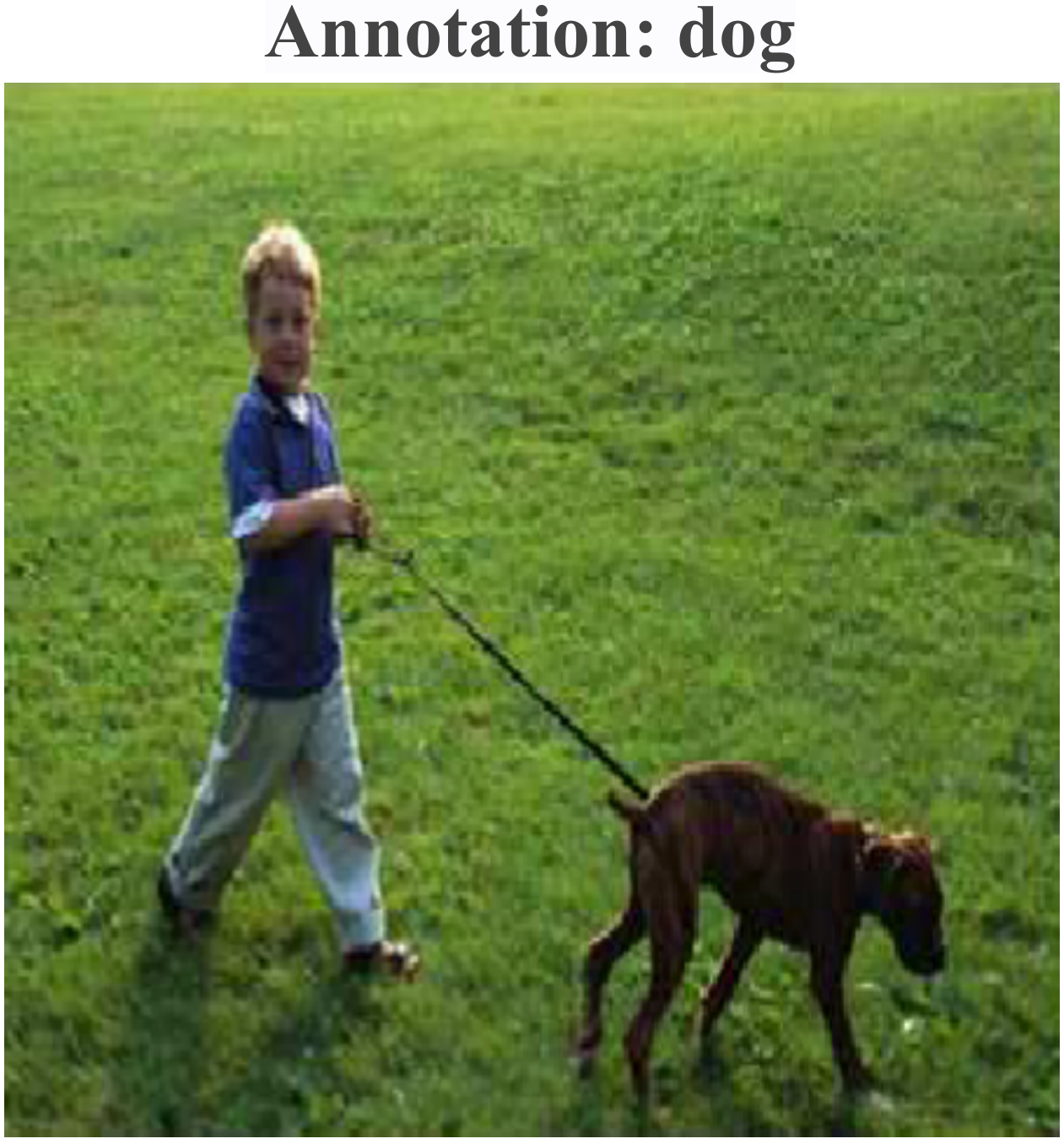}
    \caption{\emph{x-noise} \uppercase\expandafter{\romannumeral2}: background noise.}
    \label{fig:xnoise2}
  \end{subfigure}
  \caption{An illustration of \emph{y-noise} and \emph{x-noise}.}
  \label{fig:noise}
\end{figure}

\section{Preliminaries and Problem Setting}
Let $\mathcal{D}=\left\{(x_{i}, y_{i})\right\}_{i=1}^{N} $ be a dataset consisting of $N$ image-label pairs, where $x_i \in \mathbb{R}^{d}$ and $y_i\in\{1,2,...,C\}$ denote the image and the label respectively.
We can routinely deploy a ${\theta}$-parameterized classifier (e.g., a DNN) $f_{\theta}: \mathbb{R}^{d} \rightarrow \Delta^{C}$ for data fitting, where $\Delta^{C}$ is the probability simplex over $C$ classes. 
In other words, the classifier defines a probability distribution ${p_\theta}(y|{x})=p(y | f_{\theta}(x))$.
Typically, we minimize the cross-entropy loss, i.e., perform maximum likelihood estimation (MLE), to train the model: 
\begin{equation}
\min_\theta \ell(\theta; \mathcal{D})=-\frac{1}{N}\sum_{i=1}^{N}  \log \left(f_{\theta}\left(x_{i}\right)[y_i]\right),
\end{equation}
where $f_{\theta}(x_{i})[y_i]$ refers to the $y_i$-th element of the vector $f_{\theta}(x_{i})$.
We can also augment the above objective with an L2 penalty on weights $||\theta||_2^2$ to achieve maximum a posteriori (MAP) estimation. 

To enable the characterization of uncertainty, Bayesian neural networks (BNNs) place a prior distribution over DNNs weights $p(\theta)$, and perform Bayesian inference to find the posterior distribution $p(\theta|\mathcal{D})$ instead of performing MLE or MAP estimation as in the deterministic DNNs. 
Such an uncertainty-aware modeling can give rise to a more calibrated predictive distribution.

\subsection{The Setting of Learning under Noise}
In practice, the collected dataset may suffer from heterogeneous \emph{noise}. 
A typical assumption on data noise is that there are systematical errors in the annotations, i.e., the label noise (\emph{y-noise}). 
For example, an image of lion may be annotated as ``tiger'' as shown in Fig.~\ref{fig:lnoise}. 
Tremendous effort has been devoted to handling
symmetric, asymmetric, or even instance-dependent \emph{y-noise}~\cite{patrini2017making,wang2019symmetric,xia2020part}.
However, in practice, the noise may exist in not only the annotations but also the images themselves (i.e., \emph{x-noise}), casting new challenges for the deep learning models in the real world.

Common \emph{x-noise} includes image corruptions~\cite{hendrycks2018benchmarking} like distortion, blur, compression, etc. (see Fig.~\ref{fig:xnoise1}).
The \emph{x-noise} may also stem from the inherent ambiguity of the image (see Fig.~\ref{fig:xnoise2}), which is termed as background noise by the previous work~\cite{tu2020learning}. 
We use \emph{x-noise} \uppercase\expandafter{\romannumeral1} and \emph{x-noise} \uppercase\expandafter{\romannumeral2} to refer to the aforementioned two types of \emph{x-noise} for short.
The \emph{x-noise} results in low-quality or even incomplete observations and may cause over-fitting and bias the model.
The existing works for dealing with \emph{x-noise} mainly focus on image corruptions (e.g., image denoising and Super-Resolution (SR)), and often require some specific assumptions~\cite{fan2019brief} or expensive computational resources~\cite{wang2020deep}. Namely, there are still barriers for them to deal with the real-world image noise~\cite{fan2019brief}, especially for the background noise.

\begin{figure}[ht]
    \centering
    \includegraphics[width=0.95\linewidth]{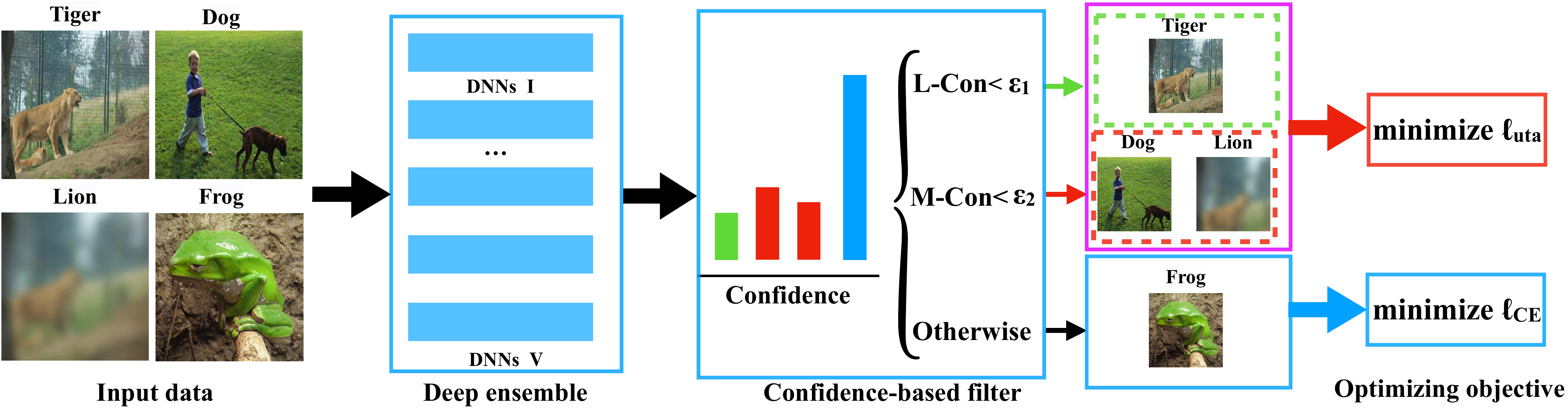}
    \caption{Overview of the proposed method. Given training data with \emph{(x,y)-noise}, the proposed method first distinguishes noisy samples from clean samples using the \emph{confidence-based sample filter}. Then, we can minimize the standard cross-entropy loss for clean data but minimize model uncertainty for noisy data in the framework. $\ell_{uta}$ represents the loss function of uncertainty penalty.}
    \label{fig:overview}
\end{figure}

In this paper, we focus on the learning under dual noises (i.e., the joint \emph{(x,y)-noise}), a more general and more challenging setting than learning under only \emph{x-noise}~\cite{hendrycks2018benchmarking,tu2020learning} or \emph{y-noise}~\cite{patrini2017making,xia2020part}. 
A naive combination of the two lines of works would suffer from the limitations in both sides, and miss the opportunities to handle x-noise and y-noise in parallel.
To address this challenge, we need to develop a unified and reliable learning strategy to avoid detrimental over-fitting.


\section{Methodology}
Uncertainty-based deep models can potentially mitigate the over-fitting to noisy data due to the inherent characterization of uncertainty.
We have conducted a thorough empirical study on using uncertainty-based deep models like BNNs and deep ensemble to handle dual noises (see Appendix \ref{appendix:deep}).
We found that uncertainty-based models can better alleviate over-fitting than deterministic DNNs. 
However, these models can still suffer from the bias in the noisy data and yield compromising results. 
As a result, we propose two strategies to further promote the effectiveness of uncertainty-based deep models for handling dual noises. 
We place our following discussion upon deep ensemble, one of the best uncertain-based deep models revealed by pioneering works~\cite{ovadia2019can} and our study, and clarify that our strategies are compatible with other backbones like BNNs.

We first briefly review deep ensemble. 
Concretely, a deep ensemble consists of $M$ randomly initialized, individually trained DNNs $\{f_{\theta_m}\}_{m=1}^M$, and makes predictions by uniform voting: 
\begin{equation}
\label{eqn:deepens}
     \frac{1}{M} \sum_{m=1}^{M} f_{\theta_{m}}(x).
\end{equation}
As shown, deep ensemble is easy to implement and flexible, which makes our approach enjoys good practicability and scalability.

In the following, we discuss how to construct a confidence-based sample filter to progressively filter out noisy samples, and how to excavate valuable information from detected noisy data. 
We illustrate our method in Fig.~\ref{fig:overview}.
\subsection{The Confidence-based Sample Filter}
Distinct from leveraging complicated strategies for noise detection in previous works~\cite{arazo2019unsupervised,liu2020peer}, we propose a simple confidence-based sample filter to filter out \emph{x-noise} and \emph{y-noise} in parallel. 

\textbf{Filtering out \emph{y-noise} using the Label confidence (L-Con).} 
Specifically, we first use the predictive probability corresponding to the label $y$ (i.e., the label confidence) to distinguish the data with \emph{y-noise} from the others.
In the case of deep ensemble, the label confidence can be simply estimated by:
\begin{equation}
     \textbf{L-Con}(x) =\frac{1}{M} \sum_{m=1}^{M} f_{\theta_{m}}(x)[y].
\label{eqn:l_con}
\end{equation}

Intuitively, \textbf{L-Con} reflects how confident the model is for the current input w.r.t. the label. 
Our hypothesis is that our model tends to yield low $\textbf{L-Con}$ for the training data with \emph{y-noise} yet yield high $\textbf{L-Con}$ for the others. 
We empirically corroborate this in Fig.~\ref{fig:lcon_dis} in Appendix. 
As shown, the data with \emph{y-noise} can be accurately distinguished from the clean data by \textbf{L-Con}. 
More importantly, the \textbf{L-Con} of the data with \emph{y-noise} is not mixed up with that of the clean data even at the later training phase (see Fig.~\ref{fig:lcon100}). 

\textbf{Filtering out \emph{x-noise} using the Maximum confidence (M-Con).} We then move to the detection of \emph{x-noise}. 
Inspired by the success of using the maximum confidence for out-of-distribution detection~\cite{hendrycks2016baseline}, we utilize the maximum confidence to detect the training data with \emph{x-noise}. 
The maximum confidence of deep ensemble takes the form of
\begin{equation}
     \textbf{M-Con}(x) = \max_j \Bigg(\frac{1}{M} \sum_{m=1}^{M} f_{\theta_{m}}(x)\Bigg)[j].
     \label{eqn:m_con}
\end{equation}

\textbf{Why M-Con is effective in detecting \emph{x-noise}?} In fact, there is an inherent connection between \textbf{M-Con} and data uncertainty (i.e., aleatoric uncertainty). Recalling the explanation in \cite{Kendall,depeweg2018decomposition}, the data/aleatoric uncertainty represents the magnitude of the inherent data noise (e.g., sensor noise), and can be estimated by 
\begin{equation}
\nonumber
    \text{Ale}(x) = \E_{p_{(\theta|\mathcal{D})}}(\mathcal{H}(p(y|{x,\theta})) = \frac{1}{M} \sum_{m=1}^{M}\mathcal{H}(p_{\theta_{m}}(y|x,\theta)),
\end{equation}
where $\mathcal{H}$ is the Shannon entropy, and it can be directly estimated by
\begin{equation}
\mathcal{H}[p(y|x)]=-\sum_{c=1}^{C} (f_{\theta}(x)[c]) (\log f_{\theta}(x)[c]),
\end{equation}
where $C$ is the number of classes. 
When the model is confident in its prediction (i.e., \textbf{M-Con} is high), it yields a sharp predictive distribution centered on one of the corners of the simplex. In contrast, when the model is not confident in its prediction (i.e., \textbf{M-Con} is low), it yields a flat predictive distribution scattered in every direction of the simplex, which corresponds to a high data uncertainty. 
There is evidence showing that the data uncertainty grows as the quality of the input image degrades~\cite{chang2020data}, so \textbf{M-Con} is effective in detecting the noisy data with \emph{x-noise}. 

Besides, \textbf{M-Con} is effective to detect the data with underlying complexity or bias. As shown in Fig.~\ref{fig:bias_data} in Appendix, some samples with low  \textbf{M-Con} correspond to the data with underlying complexity or bias (i.e., hard or dirty samples). As evidenced by some closely related works \cite{mindermann2022prioritized,chang2020data}, detecting the data with underlying complexity (bias) for a particular treatment can improve the model's predictive performance or training efficiency.

\textbf{How to filter?} We propose a simple yet efficient sample filter based on \textbf{L-Con} and \textbf{M-Con}. To be specific, we first assign different weights for different data according to the value of \textbf{L-Con},
\begin{equation}
    w^{l}_{i} = \begin{cases}0, & \text { if } \textbf{L-Con}(x_i) \leq \epsilon_1 \\ 1, & \text{otherwise,} \end{cases}
    \label{eqn:l_thres}
\end{equation}
where $\epsilon_1$ is the threshold for filtering out \emph{y-noise}, and $w^{l}_{i}$ indicates whether the label of input sample is noisy ($w^{l}_{i}$ = 0) or clean ($w^{l}_{i}$ = 1). Likewise, we can also filter out the samples with \emph{x-noise} according to the value of \textbf{M-Con}:
\begin{equation}
    w^{k}_{i} = \begin{cases}0, & \text { if } \textbf{M-Con}(x_i) \leq \epsilon_2 \\ 1, & \text{otherwise.} \end{cases}
    \label{eqn:m_thres}
\end{equation}
$\epsilon_2$ is the threshold to decide whether the input sample is clean ($w^{k}_{i}$ = 1) or not ($w^{k}_{i}$ = 0). 

After twice filtering, the final sample weight is $w_i^s = w^{l}_{i} \times w^{k}_{i}$.
Generally, we first train the deep ensemble under a high learning rate for some epochs, after which we use the confidence-based sample filter to filter out noisy data at per iteration. The foregoing warm-up can make the sample filter better for distinguish the noisy data from the clean one.

Furthermore, we perform quantitative experiments to demonstrate the efficacy of the filters for detecting noisy data with \emph{x-noise} and \emph{y-noise}. Concretely, we regard the detection of noisy data as a binary classification problem and use the Area Under the Receiver Operating Characteristic curve (AUROC) to indicate the effectiveness of our filter. As shown in Table \ref{tab:auroc} in Appendix, the confidence-based (i.e., \textbf{M-Con} and \textbf{L-Con}) sample filter can achieve high AUROCs.
\subsection{Uncertainty Penalty on Noisy Data}
We first discuss the limitations of the typical learning objectives for dealing with dual noises. 
Then, we propose an improved learning objective based on model uncertainty.

\textbf{Limitations of typical learning objectives.}
After distinguishing the clean samples from the noisy ones, it is necessary to resort to some new learning objectives to drive the model training, since that continuing pushing the model to fit dual noises may exacerbate the over-fitting. 
Typical strategies like the loss correction technique~\cite{reed2014training,arazo2019unsupervised} regard the model predictions as pseudo labels and minimize the following loss
\begin{equation}
    \ell({\theta;\mathcal{D}})=-\sum_{i=1}^{N}\Bigg(\alpha_i \log \left(f_{\theta}\left(x_{i}\right)[y_i]\right) + \beta_i \sum_{c=1}^{C} \mathtt{stop\_grad}(f_{\theta}(x_{i})[c]) \log(f_{\theta}(x_{i})[c])\Bigg) ,
\end{equation}
where $\alpha$ and $\beta$ are the weights for clean data and noisy labels. 

Nevertheless, it is non-trivial to extent these strategies to dealing with the data with \emph{x-noise}. 
The model cannot make reliable predictions for the images with \emph{x-noise}, so taking them as pseudo labels may be harmful. 

\textbf{The model uncertainty estimation.} Fortunately, we notice that deep ensemble can offer high-quality measures of model uncertainty for the input data~\cite{Lakshminarayanan,ovadia2019can}. By penalizing the model uncertainty of noisy data, we can make our model certain on the training data with \emph{(x,y)-noise}. Specifically, the model uncertainty can be measured by the mutual information between the predictions and the model parameters~\cite{depeweg2018decomposition,smith2018understanding}.
\begin{equation}
\small
\nonumber
\label{eq_mi}
\underbrace{\mathcal{I}\left[y, \theta | {x}; \mathcal{D}\right]}_{\text {Model Uncertainty }}=\underbrace{\mathcal{H}\left[\E_{p_{(\theta|\mathcal{D})}}(p(y|x,\theta))\right]}_{\text {Total Uncertainty }}-\underbrace{\E_{p_{(\theta|\mathcal{D})}}[\mathcal{H}(p(y|{x,\theta})]}_{\text {Data Uncertainty }},
\end{equation}
which, in the case of deep ensemble, boils down to 
\begin{equation}
    \mathcal{I}\left[y, \theta | {x}; \mathcal{D}\right] \approx \mathcal{H}[\frac{1}{M} \sum_{m=1}^{M}p_{\theta_{m}}(y|x)] 
    - \frac{1}{M} \sum_{m=1}^{M}\mathcal{H}[p_{\theta_{m}}(y|x)].
\end{equation}

\textbf{The proposed learning objective.} 
Specifically, we optimize the following loss for each ensemble member in deep ensemble:
\begin{equation}
\label{lossfunc}
    \min_{\theta_{m}} \ell(\theta_m; \mathcal{D})= \begin{cases}\sum_{i=1}^{N} -\log \left(f_{\theta_m}\left(x_{i}\right)[y_i]\right), & \text {if } w^{s}_{i} = 1 \\ \sum_{i=1}^{N}\mathcal{I}(y, \theta|x_i,\mathcal{D}), & \text{if } w^{s}_{i}=0 \end{cases}
\end{equation}
where $w^{s}_{i}$ is the weight of each sample. 
Namely, we minimize the standard cross-entropy loss for clean data, but minimize the model uncertainty for noisy data. 
Intuitively, the former allows the model to constantly learn useful information when the labels and images are reliable. 
The latter enables the model to explore the valuable information inside the noisy data, while preventing the model from being misled by the harmful supervisory information.
We detail the whole process of the proposed method in Algorithm \ref{alg:main}.
\newcommand\mycommfont[1]{\footnotesize\ttfamily{#1}}
\SetCommentSty{mycommfont}

\begin{algorithm}
\caption{Training DNNs under \emph{(x,y)-noise}}\label{alg:main}
\KwIn{Training noisy dataset $\mathcal D$, number of networks $M$ for ensemble, \textbf{L-Con} threshold $\epsilon_1$, \textbf{M-Con} threshold $\epsilon_2$}

Initialize $M$ networks $f_{\theta_1},\cdots,f_{\theta_M}$\;
\For {$m=1:M$} {$\theta^{(m)}\leftarrow\operatorname{WarmUp}(\mathcal D, \theta^{(m)})$\;
} 

\While{$e<\operatorname{MaxEpoch}$}{
    \For {Mini-batch $\mathcal B$ in $\mathcal D$}{
        Compute \textbf{L-Con} and \textbf{M-Con} using equation \ref{eqn:l_con} and \ref{eqn:m_con}\;
        Determine weights $w_i^l$ and $w_i^k$ following thresholding rule \ref{eqn:l_thres} and \ref{eqn:m_thres}\;
        Update each network $f_{\theta_m}$ with loss function $\mathcal L(\theta_m,\mathcal B)=\sum_{(x_i,y_i)\in \mathcal B}(1-w_i^kw_i^l)\mathcal I(y_i, \theta)+w_i^kw_i^l\mathcal{L}_{\text{CE}}(\theta_m,\mathcal B)$\;
    } 
    $e=e+1$
}
\end{algorithm}

\section{Experiment}
In this section, we first evaluate the proposed method on datasets with synthetic noise and the real-world dataset WebVision. Furthermore, we ablate the robustness of the proposed method to hyper-parameters in terms of the number of ensembles: $M$ and two thresholds: $\epsilon_1$ and $\epsilon_2$. We also verify the effectiveness of the uncertainty penalty strategy in ablation studies.

\textbf{Datasets.} The proposed method is first evaluated on two benchmark datasets with synthetic noise: CIFAR-100~\cite{krizhevsky2009learning} and TinyImageNet~\cite{krizhevsky2009learning} (the subset of ImageNet\cite{deng2009imagenet}), the former consists of 100 classes with 32x32 color images, and the latter has 200 classes with 64x64 color images. Moreover, we validate the effectiveness of the proposed method under more challenging real-world noise on WebVision~\cite{li2017webvision}, which contains more than 2.4 million images crawled from the Flickr website and Google Images search.

\textbf{Implementation details.} The synthetic noise contains the common \emph{y-noise} used in \cite{zhang2017understanding,arazo2019unsupervised} and \emph{x-noise} \uppercase\expandafter{\romannumeral1}: the corruption on images. We use the symmetric noise as the synthetic \emph{y-noise}, which is generated by randomly flipping the true label to other possible labels. For \emph{x-noise} \uppercase\expandafter{\romannumeral1}, we randomly apply the challenging ``Gaussian Blur'', ``Fog'' and ``Contrast'' corruption used in \cite{hendrycks2018benchmarking} to the original images to simulate the real-world image noise. The \emph{x-noise} \uppercase\expandafter{\romannumeral2} (i.e., background noise) commonly exists in web images, thus we also evaluate the proposed method on WebVision dataset. The deep ensemble we used consists of 5 ResNet18~\cite{he2016deep} for all datasets. SGD is used to optimize the network with a batch size of 256. More details can be found in Appendix~\ref{appendix:exp}.

\textbf{Baselines.} The first thing to note is that all methods employ 5 networks for fair comparisons. We compare with two kinds of compared baselines. The first kind contains the single model (Single-CE) and deep ensemble (DE-CE) with the standard cross-entropy loss. The second kind is competitive loss correction technique related to our method, which involves the regularized loss function with dynamic bootstrapping (DYR)~\cite{arazo2019unsupervised}, the regularized loss function with mixup dynamic bootstrapping (M-DYR)~\cite{arazo2019unsupervised} and COnfidence REgularized Sample Sieve (CORES$^2$)~\cite{cheng2020learning}. Besides, we use ``Proposed-L (Proposed-M)'' to indicate that we only use \textbf{L-Con} (\textbf{M-Con}) to filter out noisy samples and use ``Proposed-LM'' to represent the proposed method with \textbf{L-Con} and \textbf{M-Con} filter. Furthermore, we also consider the pipeline of combining the denoising technique and M-DYR as a compared baseline. However, as shown in Fig.~\ref{deblur} in Appendix, we can observe that existing denoising methods do not restore globally blurred images. As a consequence, a more effective strategy is to filter out low-quality images in this paper rather than restore them.

\vspace*{-0.25cm}
\subsection{Performance under Synthetic {(x,y)-noise}}
In this section, we first empirically evaluate the proposed method and other baselines on CIFAR-100 and TinyImageNet with different levels of synthetic \emph{(x,y)-noise}. Afterward, we also compare the proposed method with competitive baselines under the label noise.

\begin{table}[ht]
\renewcommand\arraystretch{1.2}
\centering
\setlength{\tabcolsep}{2.2mm}
\caption{The comparison of validation accuracy on CIFAR-100 and TinyImageNet with \emph{(x,y)-noise}. ``$0.2y+0.3x$'' represents the dataset with 20\% \emph{y-noise} and 30\% \emph{x-noise} simultaneously.}
\begin{tabular}{@{}lcccccc@{}}
\toprule
Alg./Noise rate &  &0.0 & 0.3$x$ & 0.4$x$ & 0.2$y$+0.3$x$ & 0.4$y$+0.3$x$ \\ \midrule
 & & \multicolumn{5}{c}{CIFAR-100 / TinyImageNet }   \\
\cline{3-7}
\multirow{1}{*} Single-CE      
& Best &77.23/61.19   &73.62/54.39   &72.53/52.53      &57.84/43.59   &47.76/40.62    \\
& Last &76.44/60.26   &72.19/49.03   &71.95/49.95      &57.39/36.81   &41.39/22.62 \\
\hline
\multirow{1}{*} DE-CE       
& Best & 79.13/63.62  &77.07/60.03   &76.12/59.94      &66.50/50.03    &54.90/46.36    \\
& Last & 77.01/61.28  &76.14/59.51   &74.98/59.05      &65.24/46.21    &53.91/41.27 \\
 \hline
\multirow{1}{*} DYR~\cite{arazo2019unsupervised}  
&Best & 78.64/\textbf{65.14}  &73.64/60.74 &71.68/59.20    &62.54/52.14     &50.54/43.94   \\
&Last & 78.02/63.97  &73.07/59.25    &71.13/58.01    &60.59/50.67     &49.21/40.89 \\
\hline
\multirow{1}{*} M-DYR~\cite{arazo2019unsupervised}   
& Best & 75.38/62.32  &75.11/60.70   &73.86/59.45    &72.38/52.14     &64.07/50.50     \\
& Last & 74.91/61.04  &74.28/58.44   &72.41/57.21    &70.69/50.04     &62.34/48.02 \\
\hline
\multirow{1}{*} CORES$^2$~\cite{cheng2020learning}            
& Best &76.76/59.74  &73.15/57.22 &72.04/55.67  &63.06/46.40 &51.98/44.55  \\
& Last &76.22/59.14  &73.01/56.35 &71.98/54.41  &62.51/44.91 &51.11/43.20 \\
\hline
\multirow{1}{*}Proposed-L 
& Best & 80.44/64.07  &77.01/60.68 &76.06/59.54  &71.05/57.62    &63.04/51.41     \\
& Last & 79.03/63.21  &76.58/59.99 &75.08/58.62  &69.91/56.31    &61.97/50.23 \\
\hline
\multirow{1}{*}Proposed-M 
& Best & 80.61/64.37  &77.89/\textbf{61.06} &77.51/\textbf{60.51}  &-/-    &-/-     \\
& Last & 79.39/64.01  &77.02/\textbf{60.26} &77.19/\textbf{59.34}  &-/-    &-/-     \\
\hline
\multirow{1}{*}Proposed-LM
&Best &\textbf{80.98}/64.58  &\textbf{77.92}/61.01   &\textbf{77.53}/60.12    &\textbf{72.78}/\textbf{58.75}    &\textbf{66.61}/\textbf{52.35}    \\
& Last & \textbf{79.71}/\textbf{64.15}  &\textbf{77.03}/60.23   &\textbf{77.32}/59.19    &\textbf{72.48}/\textbf{57.82}    &\textbf{66.05}/\textbf{51.02} \\ 
\bottomrule
\end{tabular}
\label{tab:noisy_xy_c100}
\end{table}

We evaluate the classification accuracy at the best and last epoch following the setting of \cite{arazo2019unsupervised}. Table~\ref{tab:noisy_xy_c100} presents the results of all methods on CIFAR-100 and TinyImageNet with different rates of \emph{x-noise} and \emph{y-noise}. We can see that the proposed method outperforms other baselines under synthetic \emph{(x,y)-noise} in terms of classification accuracy at the best and last epoch. Especially, the proposed method achieves a remarkable performance improvement comparing other methods under the joint \emph{(x,y)-noise} (i.e., ``0.2$y$+0.3$x$'' and ``0.4$y$+0.3$x$'' in Table~\ref{tab:noisy_xy_c100}), which shows the effectiveness of the proposed method to handle dual noises. 

Besides, we can observe that Proposed-M outperforms DE-CE under \emph{x-noise} (i.e, ``0.3$x$'' and ``0.4$x$''), which shows that the effectiveness of employing \textbf{M-Con} to filter out samples with \emph{x-noise}. By contrast, the previous works that focus on the noisy label (i.e., DYR, M-DYR and CORES$^2$) do not show the superior performance regardless of whether \emph{x-noise} or \emph{(x,y)-noise}, which confirms that they cannot effectively handle dual noises.
Moreover, we can notice that the naive deep ensemble with cross-entropy loss (DE-CE) significantly outperforms the single model (Single-CE), confirming that uncertainty-based deep ensemble can prevent the model from over-fitting noisy data. In addition, we can notice that the experimental results exhibit quite close best accuracy and last accuracy, which shows that our method is not easy to over-fit noisy data and can achieve stable and robust learning.

To verify the effectiveness of the proposed method under label noise (i.e., \emph{y-noise}), we also compare our method with other baselines on CIFAR-100 and TinyImageNet with different levels of synthetic \emph{y-noise} in Appendix~\ref{appendix-y}. The experimental results show that the proposed method significantly outperforms competitive methods for noisy labels. Specifically, ``Proposed-L'' also outperforms or is close to the best results of other baselines. We discuss more details in Appendix~\ref{appendix-y}.
\begin{table}[!ht]
\centering
\setlength{\tabcolsep}{1.1mm}
\caption{The comparison of validation accuracy on ImageNet ILSVRC12 and WebVision validation set. The number outside (inside) the parentheses
denotes top-1 (top-5) accuracy.}
\begin{tabular}{@{}lcccccc@{}}
\toprule
Val./Methods  & &DYR &M-DYR &CORES$^2$ &DE-CE &Proposed-LM \\
\midrule
\multirow{1}{*}WebVision           
&Best &69.48 (83.21) &72.36 (87.40) &70.56 (87.56) &73.76 (88.13) &\textbf{76.68 (91.32)} \\
&Last &68.53 (82.42)  &72.01 (87.15) &69.52 (87.02) &73.22 (87.98) &\textbf{76.52 (91.22)} \\
\hline
\multirow{1}{*}ILSVRC12   
& Best &67.32 (89.76) & 68.52 (86.36)  & 64.12 (86.36) &67.64 (88.73) &\textbf{71.40 (90.88)}\\
& Last &66.59 (88.98) & 68.33 (86.21)  & 63.23 (85.44) &67.31 (88.26)  &\textbf{71.26 (90.70)}\\
\bottomrule
\end{tabular}
\label{tab:webvision}
\end{table}
\subsection{Performance on the Real-world Noisy Dataset}
Furthermore, we verify the generalization performance of the proposed method on a large real-world noisy dataset: WebVision. Since the dataset is too big, for quick experiments, we compare all methods on the first 50 classes (denoted as WebVision-50) of the Google image subset and use the resized images following previous works~\cite{jiang2018mentornet,chen2019understanding}. Besides, we test the trained model of all methods on the human-annotated WebVision validation set and the ILSVRC12 validation set~\cite{deng2009imagenet}. Table \ref{tab:webvision} lists the experimental results. As we can see, the proposed method significantly outperforms other baselines not only on the WebVision validation set but also on the ILSVRC12 validation set for the real-world noisy dataset, which shows the superiority of our method is also effective to the real-world noisy dataset. 

\begin{figure}
  \centering
  \begin{subfigure}{0.50\linewidth}
    \includegraphics[width=0.98\linewidth]{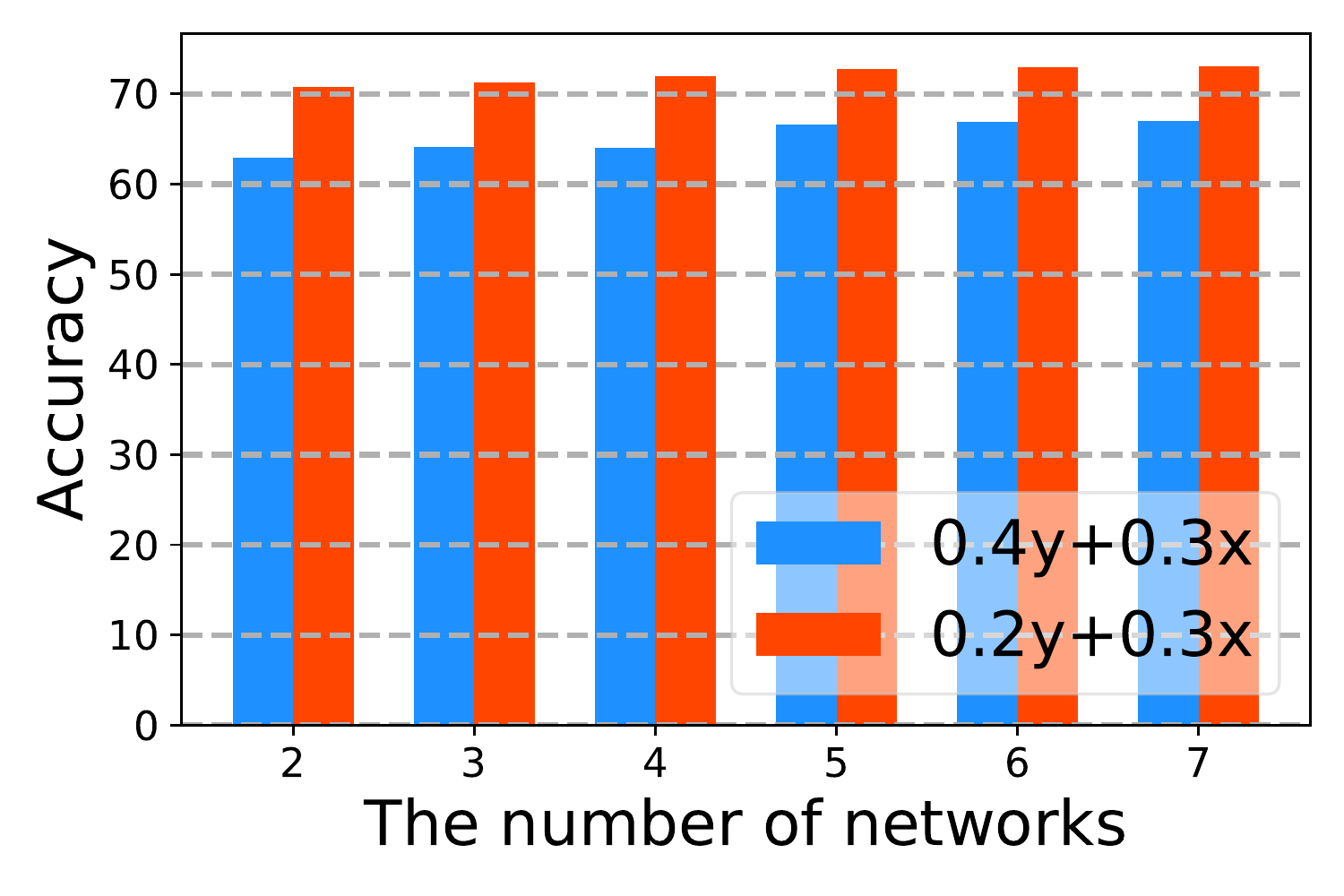}
    \caption{The performance under different levels of (x,y)-noise.}
    \label{fig:ensemble_xy}
  \end{subfigure}
  \hfill
  \begin{subfigure}{0.48\linewidth}
    \includegraphics[width=0.98\linewidth]{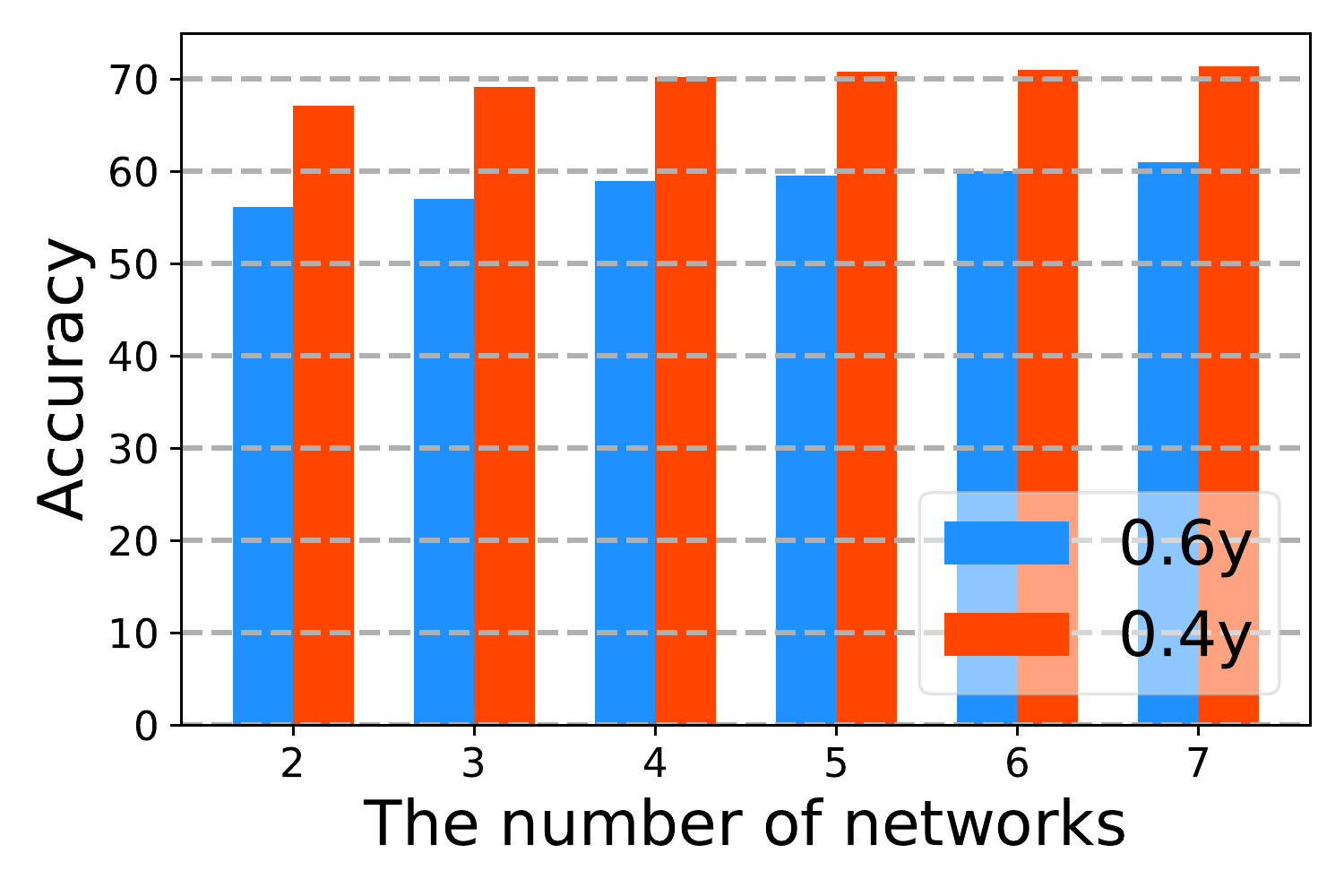}
    \caption{The performance under different levels of y-noise.}
    
    \label{fig:ensemble_y}
  \end{subfigure}
  \caption{Effects of different numbers of networks on the performance of the proposed method on CIFAR-100.}
  \label{fig:ab-ensemble}
\end{figure}
\subsection{Ablation Studies}

\textbf{Empirical effects of the number of networks $M$.} The number of networks for deep ensemble is a crucial hyper-parameter. Empirically, the more networks for deep ensemble, the more powerful performance can achieve. However, assembling a large number of networks often requires high memory and computational costs. Hence, we need to make an appropriate trade-off between the performance and the computational cost. Fig.~\ref{fig:ab-ensemble} demonstrates the performance (i.e., the best validation accuracy) of the proposed method corresponding to the different numbers of networks under different levels of \emph{(x,y)-noise} on CIFAR-100. We can see that even a small number of networks can not overly drop the performance. When the number of networks is greater than 4, the proposed method can almost achieve the best performance, so an ensemble of 5 networks is enough for our method.

\textbf{Empirical effects of thresholds of confidence-based sample filter.} Moreover, we analyze the effects of hyper-parameters: $\epsilon_1$ and $\epsilon_2$ of the proposed confidence-based sample filter on the predictive performance. For the threshold of \textbf{M-Con}, we use a soft threshold to filter out the training data with \emph{x-noise} after per iteration (i.e., the training data with minimum $\epsilon_2\%$ \textbf{M-Con} is filtered out), which is more effective than the hard threshold through empirical studies. For the threshold of \textbf{L-Con}, the hard threshold is more appropriate according to the empirical results in Fig.~\ref{fig:lcon_dis}. Table \ref{tab:ab_thre} reports the comparison results of different thresholds on CIFAR-100. We can observe that the performance of the proposed method is not sensitive to $\epsilon_1$ and $\epsilon_2$, which can achieve superior performance within a certain range of thresholds. Especially, all results of the proposed method are better than the baselines under the joint \emph{(x,y)-noise} in Table \ref{tab:ab_thre}. In summary, our method indeed shows the effectiveness and practicability of dealing with noisy data, which does not rely on time-consuming hyper-parameters tuning.
\begin{table}[!ht]
\centering
\setlength{\tabcolsep}{1.0mm}
\caption{The comparison of validation accuracy under different $\epsilon_1$ and $\epsilon_2$ on CIFAR-100 with different levels of \emph{(x,y)-noise}.}
\begin{tabular}{@{}lcccccc|cccccc@{}}
\toprule
$\epsilon_1$ ($10^{-2}$)  & & 1.5 & 2.0 & 2.5 & 3.0 &3.5 &$\epsilon_2$ (\%)  & 2.0 & 3.0 & 4.0 & 5.0 &6.0\\ \midrule
\multirow{1}{*}0.4$y$+0.3$x$      
&Best  &66.01 &66.61 &66.57  &66.34 &\textbf{66.58} 
&Best &64.92  &65.81   & 66.02  &\textbf{66.72}   & 66.11 \\
&Last  &65.63 &66.05 &66.04  &\textbf{66.10} &66.01 
&Last &64.43  &64.52   & 65.59  &\textbf{66.09}   &66.04\\
\hline
\multirow{1}{*}0.2$y$+0.3$x$           
&Best  &71.92   &\textbf{72.78}    &72.61    &72.69  &72.71 
&Best &71.93   &72.51    &72.76  &\textbf{72.93} &72.88\\
&Last  &71.22   &\textbf{72.48}    &72.24    &72.45  &72.44 
&Last &71.85   &72.33    &72.24  &\textbf{72.59} &72.49\\
\hline
\multirow{1}{*}0.6$y$   
& Best &57.85   &\textbf{59.65}    &59.52    &58.59  &57.94 
& Best &58.20 &58.92    &58.99    &59.07   &\textbf{59.61}\\
& Last &55.96   &\textbf{55.53}    &55.26    &55.06  &55.03 
& Last &57.01 &56.63    &56.17    &\textbf{56.44}   &55.21\\
\hline
\multirow{1}{*}0.4$y$           
&Best &70.63   &\textbf{70.77}    &70.28    &70.11 &69.88 
&Best &69.82   &69.98    &\textbf{70.81}    &70.62 &70.59\\
&Last &68.27   &\textbf{68.83}    &68.19    &67.63 &67.46 
&Last &67.89   &68.07    &68.94    &68.71 &\textbf{68.72}\\
\bottomrule
\end{tabular}
\label{tab:ab_thre}
\end{table}

\textbf{Effects of uncertainty penalty in the proposed learning objective.} To verify the effectiveness of uncertainty penalty in Eqn.~\eqref{lossfunc}, we report the performance of the proposed method without uncertainty penalty and the gap with ``Proposed-LM'' in Table~\ref{tab:loss_abla}. We observe that the validation accuracy is lower than the complete workflow on both \emph{y-noise} and \emph{(x,y)-noise}, which clarify that the effectiveness of the uncertainty penalty strategy.

\begin{table}[ht]
\centering
\renewcommand\arraystretch{1.2}
\setlength{\tabcolsep}{1.0mm}
\caption{The best accuracy on CIFAR-100 and TinyImageNet with \emph{(x,y)}-noise.}
\begin{tabular}{@{}ccccc@{}}
\toprule
Noise rate & 0.4$y$ & 0.6$y$ & 0.2$y$+0.3$x$ & 0.4$y$+0.3$x$ \\ \midrule
 & \multicolumn{4}{c}{CIFAR-100 / TinyImageNet }   \\
\cline{2-5}
\multirow{1}{*}       
Best Acc &67.92/54.23   &57.33/41.52    &70.81/55.34  &63.02/49.68    \\
Gaps  &2.85/1.98 &2.17/3.13 &1.97/3.41 &3.59/2.67 \\
\bottomrule
\end{tabular}
\label{tab:noisy_xy_c100_1}
\label{tab:loss_abla}
\end{table}


\section{Conclusions and limitations}
\subsection{Conclusions}
This work first introduces the more challenging and closer to real-world noise setting and then performs a systematical investigation on using uncertainty-based models under dual noises (i.e., the joint \emph{(x,y)-noise}). We find that merely employing an uncertainty-based model is not enough and furthermore propose a novel workflow for the learning of uncertainty-based deep models. Concretely, we present the efficient and practical confidence-based sample filter to distinguish noisy data from clean data progressively. After doing so, we propose to penalize the model uncertainty of noisy data without reliance on the misleading supervisory information. Empirically, the proposed method significantly outperforms the competitive baselines on CIFAR-100 and TinyImageNet with synthetic \emph{(x,y)-noise} and the real-world noisy dataset. We further evaluate the robustness of hyper-parameters in our method, which shows that the proposed method is not sensitive to crucial hyper-parameters. In the future, this work may promote more approaches to deal with dual noises in more tasks.
\subsection{Potential Limitations}
Though the proposed method shows superior performance, there are also some potential limitations. First, one limitation of this work is that we have not separately handled the two kinds of \emph{x-noise} for better noise detection and model training, so designing a more fine-grained approach might be helpful for model learning, and it deserves future investigation.
Second, deep ensemble is usually computationally expensive especially when the model and data complexity is high. To address the limitation, we can use computationally cheap deep ensemble but with comparable performance, such as BatchEnsemble~\cite{wen2020batchensemble} and Hyperparameter ensembles~\cite{wenzel2020hyperparameter}, whose computational and memory costs are significantly lower than typical ensembles.
\section*{Acknowledgement}

This work was supported by National Key Research and Development Project of China (No. 2021ZD0110502), NSFC Projects (Nos. 62061136001, 62076145, 62076147, U19B2034, U1811461, U19A2081, 61972224), Beijing NSF Project (No. JQ19016), BNRist (BNR2022RC01006), Tsinghua Institute for Guo Qiang, and the High Performance Computing Center, Tsinghua University. J.Z is also supported by the XPlorer Prize.



\begin{thebibliography}{10}

\bibitem{arazo2019unsupervised}
Eric Arazo, Diego Ortego, Paul Albert, Noel O’Connor, and Kevin McGuinness.
\newblock Unsupervised label noise modeling and loss correction.
\newblock In {\em International Conference on Machine Learning}, pages
  312--321. PMLR, 2019.

\bibitem{arpit2017closer}
Devansh Arpit, Stanisaw Jastrzebski, Nicolas Ballas, David Krueger, Emmanuel
  Bengio, Maxinder~S Kanwal, Tegan Maharaj, Asja Fischer, Aaron Courville,
  Yoshua Bengio, et~al.
\newblock A closer look at memorization in deep networks.
\newblock In {\em International Conference on Machine Learning}, pages
  233--242. PMLR, 2017.

\bibitem{blundell2015weight}
Charles Blundell, Julien Cornebise, Koray Kavukcuoglu, and Daan Wierstra.
\newblock Weight uncertainty in neural networks.
\newblock In {\em Proceedings of the 32nd International Conference on
  International Conference on Machine Learning-Volume 37}, pages 1613--1622,
  2015.

\bibitem{carbajal2021single}
Guillermo Carbajal, Patricia Vitoria, Mauricio Delbracio, Pablo Mus{\'e}, and
  Jos{\'e} Lezama.
\newblock Non-uniform motion blur kernel estimation via adaptive decomposition.
\newblock {\em arXiv e-prints}, pages arXiv--2102, 2021.

\bibitem{chang2020data}
Jie Chang, Zhonghao Lan, Changmao Cheng, and Yichen Wei.
\newblock Data uncertainty learning in face recognition.
\newblock In {\em Proceedings of the IEEE/CVF Conference on Computer Vision and
  Pattern Recognition}, pages 5710--5719, 2020.

\bibitem{charoenphakdee2019symmetric}
Nontawat Charoenphakdee, Jongyeong Lee, and Masashi Sugiyama.
\newblock On symmetric losses for learning from corrupted labels.
\newblock In {\em International Conference on Machine Learning}, pages
  961--970. PMLR, 2019.

\bibitem{chen2019understanding}
Pengfei Chen, Ben~Ben Liao, Guangyong Chen, and Shengyu Zhang.
\newblock Understanding and utilizing deep neural networks trained with noisy
  labels.
\newblock In {\em International Conference on Machine Learning}, pages
  1062--1070. PMLR, 2019.

\bibitem{cheng2020learning}
Hao Cheng, Zhaowei Zhu, Xingyu Li, Yifei Gong, Xing Sun, and Yang Liu.
\newblock Learning with instance-dependent label noise: A sample sieve
  approach.
\newblock In {\em International Conference on Learning Representations}, 2020.

\bibitem{deng2009imagenet}
Jia Deng, Wei Dong, Richard Socher, Li-Jia Li, Kai Li, and Li~Fei-Fei.
\newblock Imagenet: A large-scale hierarchical image database.
\newblock In {\em 2009 IEEE conference on computer vision and pattern
  recognition}, pages 248--255. Ieee, 2009.

\bibitem{depeweg2018decomposition}
Stefan Depeweg, Jose-Miguel Hernandez-Lobato, Finale Doshi-Velez, and Steffen
  Udluft.
\newblock Decomposition of uncertainty in bayesian deep learning for efficient
  and risk-sensitive learning.
\newblock In {\em International Conference on Machine Learning}, pages
  1184--1193. PMLR, 2018.

\bibitem{dodge2016understanding}
Samuel Dodge and Lina Karam.
\newblock Understanding how image quality affects deep neural networks.
\newblock In {\em 2016 eighth international conference on quality of multimedia
  experience (QoMEX)}, pages 1--6. IEEE, 2016.

\bibitem{fan2019brief}
Linwei Fan, Fan Zhang, Hui Fan, and Caiming Zhang.
\newblock Brief review of image denoising techniques.
\newblock {\em Visual Computing for Industry, Biomedicine, and Art},
  2(1):1--12, 2019.

\bibitem{fergus2006removing}
Rob Fergus, Barun Singh, Aaron Hertzmann, Sam~T Roweis, and William~T Freeman.
\newblock Removing camera shake from a single photograph.
\newblock In {\em ACM SIGGRAPH 2006 Papers}, pages 787--794. 2006.

\bibitem{fort2019deep}
Stanislav Fort, Huiyi Hu, and Balaji Lakshminarayanan.
\newblock Deep ensembles: A loss landscape perspective.
\newblock {\em arXiv preprint arXiv:1912.02757}, 2019.

\bibitem{gal2016dropout}
Yarin Gal and Zoubin Ghahramani.
\newblock Dropout as a {B}ayesian approximation: Representing model uncertainty
  in deep learning.
\newblock In {\em international conference on machine learning}, pages
  1050--1059, 2016.

\bibitem{ghosh2017robust}
Aritra Ghosh, Himanshu Kumar, and PS~Sastry.
\newblock Robust loss functions under label noise for deep neural networks.
\newblock In {\em Proceedings of the AAAI Conference on Artificial
  Intelligence}, volume~31, 2017.

\bibitem{han2018co}
Bo~Han, Quanming Yao, Xingrui Yu, Gang Niu, Miao Xu, Weihua Hu, Ivor~W Tsang,
  and Masashi Sugiyama.
\newblock Co-teaching: Robust training of deep neural networks with extremely
  noisy labels.
\newblock In {\em NeurIPS}, 2018.

\bibitem{he2017mask}
Kaiming He, Georgia Gkioxari, Piotr Doll{\'a}r, and Ross Girshick.
\newblock Mask r-cnn.
\newblock In {\em Proceedings of the IEEE international conference on computer
  vision}, pages 2961--2969, 2017.

\bibitem{he2016deep}
Kaiming He, Xiangyu Zhang, Shaoqing Ren, and Jian Sun.
\newblock Deep residual learning for image recognition.
\newblock In {\em Proceedings of the IEEE conference on computer vision and
  pattern recognition}, pages 770--778, 2016.

\bibitem{hendrycks2018benchmarking}
Dan Hendrycks and Thomas Dietterich.
\newblock Benchmarking neural network robustness to common corruptions and
  perturbations.
\newblock In {\em International Conference on Learning Representations}, 2018.

\bibitem{hendrycks2016baseline}
Dan Hendrycks and Kevin Gimpel.
\newblock A baseline for detecting misclassified and out-of-distribution
  examples in neural networks.
\newblock In {\em International Conference on Learning Representations}, 2016.

\bibitem{jiang2018mentornet}
Lu~Jiang, Zhengyuan Zhou, Thomas Leung, Li-Jia Li, and Li~Fei-Fei.
\newblock Mentornet: Learning data-driven curriculum for very deep neural
  networks on corrupted labels.
\newblock In {\em International Conference on Machine Learning}, pages
  2304--2313. PMLR, 2018.

\bibitem{Kendall}
Alex Kendall and Yarin Gal.
\newblock What uncertainties do we need in {B}ayesian deep learning for
  computer vision?
\newblock {\em Advances in neural information processing systems}, pages
  5574--5584, 2017.

\bibitem{krizhevsky2009learning}
Alex Krizhevsky, Geoffrey Hinton, et~al.
\newblock Learning multiple layers of features from tiny images.
\newblock 2009.

\bibitem{Lakshminarayanan}
Balaji Lakshminarayanan, Alexander Pritzel, and Charles Blundell.
\newblock Simple and scalable predictive uncertainty estimation using deep
  ensembles.
\newblock {\em Advances in neural information processing systems}, pages
  6402--6413, 2017.

\bibitem{deep2015}
Yann LeCun, Yoshua Bengio, and Geoffrey Hinton.
\newblock Deep learning.
\newblock {\em nature}, 521(7553):436--444, 2015.

\bibitem{levin2009understanding}
Anat Levin, Yair Weiss, Fredo Durand, and William~T Freeman.
\newblock Understanding and evaluating blind deconvolution algorithms.
\newblock In {\em 2009 IEEE Conference on Computer Vision and Pattern
  Recognition}, pages 1964--1971. IEEE, 2009.

\bibitem{li2017webvision}
Wen Li, Limin Wang, Wei Li, Eirikur Agustsson, and Luc Van~Gool.
\newblock Webvision database: Visual learning and understanding from web data.
\newblock {\em arXiv preprint arXiv:1708.02862}, 2017.

\bibitem{liu2020simple}
Jeremiah~Zhe Liu, Zi~Lin, Shreyas Padhy, Dustin Tran, Tania Bedrax-Weiss, and
  Balaji Lakshminarayanan.
\newblock Simple and principled uncertainty estimation with deterministic deep
  learning via distance awareness.
\newblock {\em Advances in Neural Information Processing Systems}, 33, 2020.

\bibitem{liu2016stein}
Qiang Liu and Dilin Wang.
\newblock Stein variational gradient descent: A general purpose {B}ayesian
  inference algorithm.
\newblock In {\em Advances in Neural Information Processing Systems}, pages
  2378--2386, 2016.

\bibitem{liu2020peer}
Yang Liu and Hongyi Guo.
\newblock Peer loss functions: Learning from noisy labels without knowing noise
  rates.
\newblock In {\em International Conference on Machine Learning}, pages
  6226--6236. PMLR, 2020.

\bibitem{maddox2019simple}
Wesley~J Maddox, Pavel Izmailov, Timur Garipov, Dmitry~P Vetrov, and
  Andrew~Gordon Wilson.
\newblock A simple baseline for bayesian uncertainty in deep learning.
\newblock In {\em Advances in Neural Information Processing Systems}, pages
  13153--13164, 2019.

\bibitem{malinin2018predictive}
Andrey Malinin and Mark Gales.
\newblock Predictive uncertainty estimation via prior networks.
\newblock In {\em Proceedings of the 32nd International Conference on Neural
  Information Processing Systems}, pages 7047--7058, 2018.

\bibitem{mao2016image}
Xiaojiao Mao, Chunhua Shen, and Yu-Bin Yang.
\newblock Image restoration using very deep convolutional encoder-decoder
  networks with symmetric skip connections.
\newblock {\em Advances in neural information processing systems},
  29:2802--2810, 2016.

\bibitem{mindermann2022prioritized}
S{\"o}ren Mindermann, Jan~M Brauner, Muhammed~T Razzak, Mrinank Sharma, Andreas
  Kirsch, Winnie Xu, Benedikt H{\"o}ltgen, Aidan~N Gomez, Adrien Morisot,
  Sebastian Farquhar, et~al.
\newblock Prioritized training on points that are learnable, worth learning,
  and not yet learnt.
\newblock In {\em International Conference on Machine Learning}, pages
  15630--15649. PMLR, 2022.

\bibitem{natarajan2013learning}
Nagarajan Natarajan, Inderjit~S Dhillon, Pradeep~K Ravikumar, and Ambuj Tewari.
\newblock Learning with noisy labels.
\newblock {\em Advances in neural information processing systems},
  26:1196--1204, 2013.

\bibitem{ovadia2019can}
Yaniv Ovadia, Emily Fertig, Jie Ren, Zachary Nado, D~Sculley, Sebastian
  Nowozin, Joshua Dillon, Balaji Lakshminarayanan, and Jasper Snoek.
\newblock Can you trust your model's uncertainty? evaluating predictive
  uncertainty under dataset shift.
\newblock {\em Advances in Neural Information Processing Systems},
  32:13991--14002, 2019.

\bibitem{patrini2017making}
Giorgio Patrini, Alessandro Rozza, Aditya Krishna~Menon, Richard Nock, and
  Lizhen Qu.
\newblock Making deep neural networks robust to label noise: A loss correction
  approach.
\newblock In {\em Proceedings of the IEEE conference on computer vision and
  pattern recognition}, pages 1944--1952, 2017.

\bibitem{reed2014training}
Scott Reed, Honglak Lee, Dragomir Anguelov, Christian Szegedy, Dumitru Erhan,
  and Andrew Rabinovich.
\newblock Training deep neural networks on noisy labels with bootstrapping.
\newblock In {\em International Conference on Learning Representations}, 2015.

\bibitem{ren2018learning}
Mengye Ren, Wenyuan Zeng, Bin Yang, and Raquel Urtasun.
\newblock Learning to reweight examples for robust deep learning.
\newblock In {\em International Conference on Machine Learning}, pages
  4334--4343. PMLR, 2018.

\bibitem{ren2015faster}
Shaoqing Ren, Kaiming He, Ross Girshick, and Jian Sun.
\newblock Faster r-cnn: Towards real-time object detection with region proposal
  networks.
\newblock {\em Advances in neural information processing systems}, 28:91--99,
  2015.

\bibitem{smith2018understanding}
Lewis Smith and Yarin Gal.
\newblock Understanding measures of uncertainty for adversarial example
  detection.
\newblock In {\em AUAI}, 2018.

\bibitem{tu2020learning}
Yi~Tu, Li~Niu, Junjie Chen, Dawei Cheng, and Liqing Zhang.
\newblock Learning from web data with self-organizing memory module.
\newblock In {\em Proceedings of the IEEE/CVF Conference on Computer Vision and
  Pattern Recognition}, pages 12846--12855, 2020.

\bibitem{van2020uncertainty}
Joost Van~Amersfoort, Lewis Smith, Yee~Whye Teh, and Yarin Gal.
\newblock Uncertainty estimation using a single deep deterministic neural
  network.
\newblock In {\em International Conference on Machine Learning}, pages
  9690--9700. PMLR, 2020.

\bibitem{wang2019symmetric}
Yisen Wang, Xingjun Ma, Zaiyi Chen, Yuan Luo, Jinfeng Yi, and James Bailey.
\newblock Symmetric cross entropy for robust learning with noisy labels.
\newblock In {\em Proceedings of the IEEE/CVF International Conference on
  Computer Vision}, pages 322--330, 2019.

\bibitem{wang2020deep}
Zhihao Wang, Jian Chen, and Steven~CH Hoi.
\newblock Deep learning for image super-resolution: A survey.
\newblock {\em IEEE transactions on pattern analysis and machine intelligence},
  2020.

\bibitem{wen2020batchensemble}
Yeming Wen, Dustin Tran, and Jimmy Ba.
\newblock Batchensemble: an alternative approach to efficient ensemble and
  lifelong learning.
\newblock {\em arXiv preprint arXiv:2002.06715}, 2020.

\bibitem{wenzel2020hyperparameter}
Florian Wenzel, Jasper Snoek, Dustin Tran, and Rodolphe Jenatton.
\newblock Hyperparameter ensembles for robustness and uncertainty
  quantification.
\newblock {\em Advances in Neural Information Processing Systems}, 33, 2020.

\bibitem{xia2020part}
Xiaobo Xia, Tongliang Liu, Bo~Han, Nannan Wang, Mingming Gong, Haifeng Liu,
  Gang Niu, Dacheng Tao, and Masashi Sugiyama.
\newblock Part-dependent label noise: Towards instance-dependent label noise.
\newblock In {\em Advances in Neural Information Processing Systems},
  volume~33, 2020.

\bibitem{xu2014deep}
Li~Xu, Jimmy~SJ Ren, Ce~Liu, and Jiaya Jia.
\newblock Deep convolutional neural network for image deconvolution.
\newblock In {\em Proceedings of the 27th International Conference on Neural
  Information Processing Systems-Volume 1}, pages 1790--1798, 2014.

\bibitem{zhang2017understanding}
Chiyuan Zhang, Samy Bengio, Moritz Hardt, Benjamin Recht, and Oriol Vinyals.
\newblock Understanding deep learning requires rethinking generalization
  (2016).
\newblock In {\em International Conference on Learning Representations}, 2017.

\bibitem{zhang2021understanding}
Chiyuan Zhang, Samy Bengio, Moritz Hardt, Benjamin Recht, and Oriol Vinyals.
\newblock Understanding deep learning (still) requires rethinking
  generalization.
\newblock In {\em International Conference on Learning Representations}, 2019.

\bibitem{zhou2017classification}
Yiren Zhou, Sibo Song, and Ngai-Man Cheung.
\newblock On classification of distorted images with deep convolutional neural
  networks.
\newblock In {\em 2017 IEEE International Conference on Acoustics, Speech and
  Signal Processing (ICASSP)}, pages 1213--1217. IEEE, 2017.

\end{thebibliography}

\section*{Checklist}

\begin{enumerate}

\item For all authors...
\begin{enumerate}
  \item Do the main claims made in the abstract and introduction accurately reflect the paper's contributions and scope?
    \answerYes{}
  \item Did you describe the limitations of your work?
    \answerYes{}
  \item Did you discuss any potential negative societal impacts of your work?
    \answerNo{We have not yet found the potential negative impact of our work.}
  \item Have you read the ethics review guidelines and ensured that your paper conforms to them?
    \answerYes{}
\end{enumerate}

\item If you are including theoretical results...
\begin{enumerate}
  \item Did you state the full set of assumptions of all theoretical results?
    \answerNA{}
	\item Did you include complete proofs of all theoretical results?
    \answerNA{}
\end{enumerate}

\item If you ran experiments...
\begin{enumerate}
  \item Did you include the code, data, and instructions needed to reproduce the main experimental results (either in the supplemental material or as a URL)?
    \answerYes{}
  \item Did you specify all the training details (e.g., data splits, hyperparameters, how they were chosen)?
    \answerYes{}
	\item Did you report error bars (e.g., with respect to the random seed after running experiments multiple times)?
    \answerYes{}
	\item Did you include the total amount of compute and the type of resources used (e.g., type of GPUs, internal cluster, or cloud provider)?
    \answerYes{}
\end{enumerate}

\item If you are using existing assets (e.g., code, data, models) or curating/releasing new assets...
\begin{enumerate}
  \item If your work uses existing assets, did you cite the creators?
    \answerYes{}
  \item Did you mention the license of the assets?  
    \answerNo{The official websites containing the datasets have already provided the corresponding license.}
  \item Did you include any new assets either in the supplemental material or as a URL?
    \answerYes{}
  \item Did you discuss whether and how consent was obtained from people whose data you're using/curating?
    \answerNo{The datasets used are all public in this paper.}
  \item Did you discuss whether the data you are using/curating contains personally identifiable information or offensive content?
    \answerNA{}
\end{enumerate}

\item If you used crowdsourcing or conducted research with human subjects...
\begin{enumerate}
  \item Did you include the full text of instructions given to participants and screenshots, if applicable?
    \answerNA{}{}
  \item Did you describe any potential participant risks, with links to Institutional Review Board (IRB) approvals, if applicable?
    \answerNA{}{}
  \item Did you include the estimated hourly wage paid to participants and the total amount spent on participant compensation?
    \answerNA{}{}
\end{enumerate}

\end{enumerate}

\newpage

\appendix

\begin{figure}[ht]
  \centering
  \begin{subfigure}{0.47\linewidth}
    \includegraphics[width=0.98\linewidth]{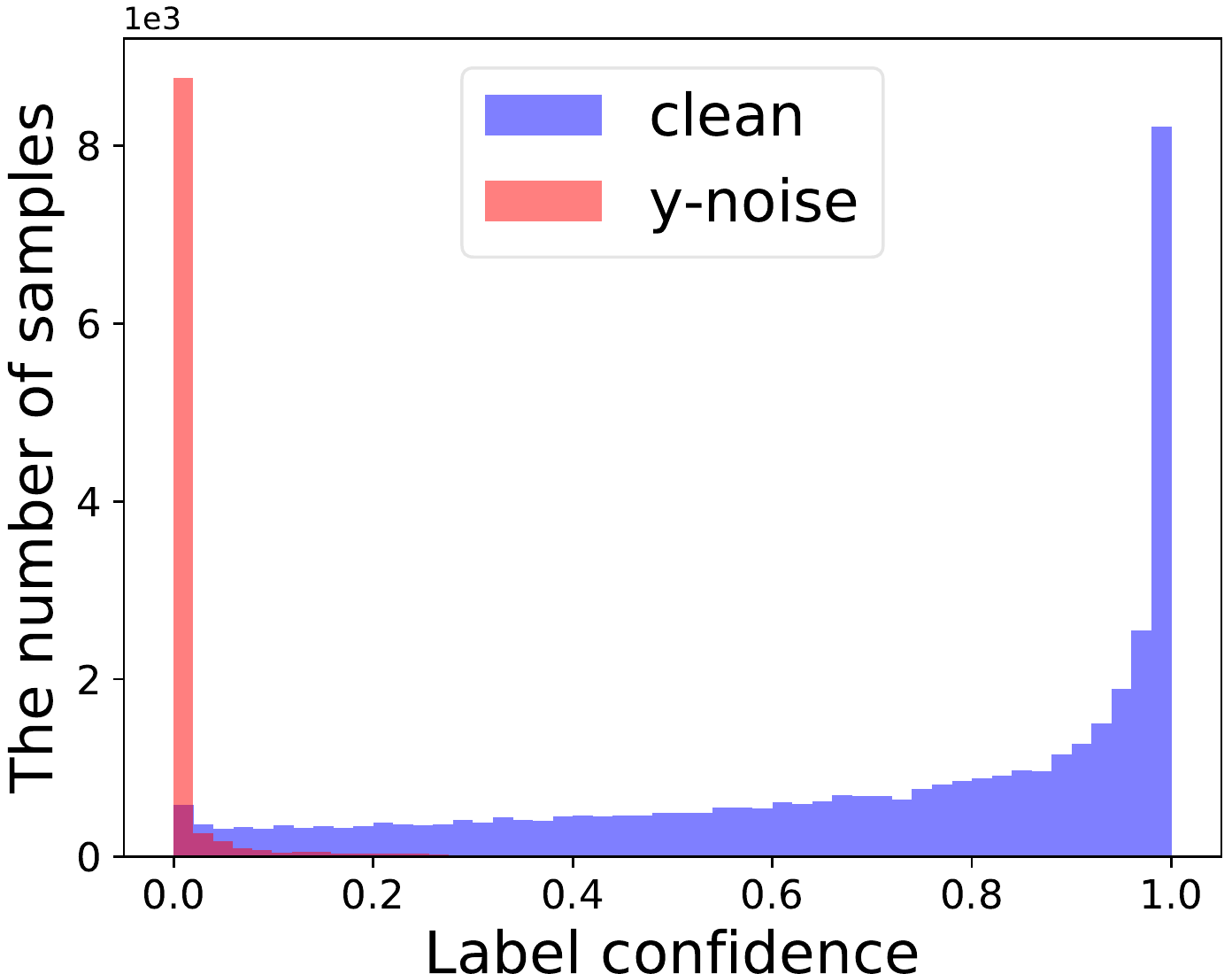}
    \caption{Histogram of \textbf{L-Con} at 40th epoch.}
    \label{fig:lcon40}
  \end{subfigure}
  \hfill
  \begin{subfigure}{0.47\linewidth}
    \includegraphics[width=0.98\linewidth]{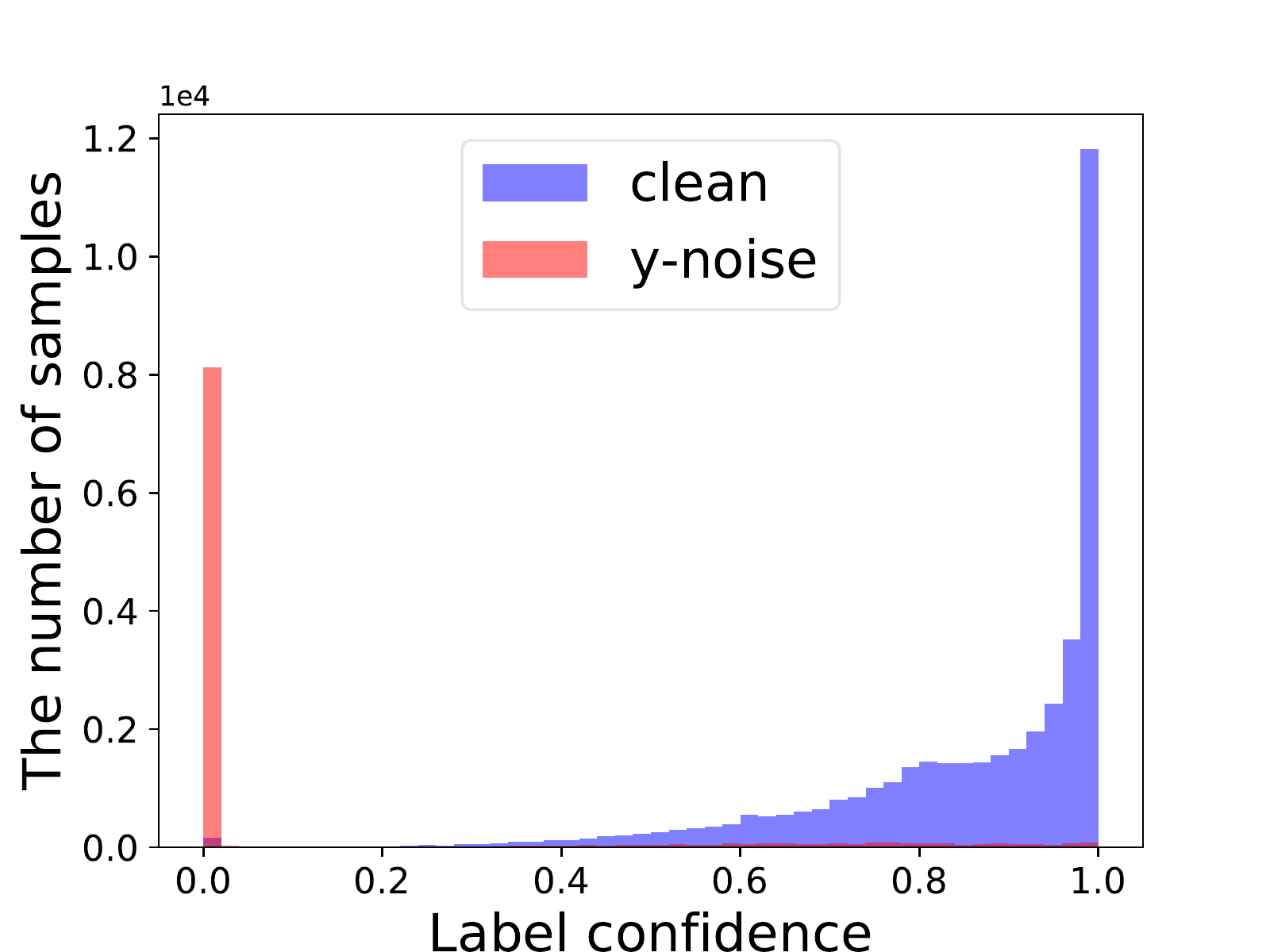}
    \caption{Histogram of \textbf{L-Con} at 100th epoch.}
    \label{fig:lcon100}
  \end{subfigure}
  \caption{\textbf{L-Con} distributions at training phase on CIFAR-100 with 40\% \emph{y-noise}.}
  \label{fig:lcon_dis}
\end{figure}
\section{The performance of different uncertainty-based models under dual noises}
\label{appendix:deep}
\begin{table}[!ht]
\centering
\setlength{\tabcolsep}{2.5mm}
\caption{The comparison of validation accuracy on CIFAR-100 with the joint \emph{(x,y)-noise}. ``$0.2y+0.3x$'' represents the dataset with 20\% \emph{y-noise} and 30\% \emph{x-noise} simultaneously.}
\begin{tabular}{@{}lccc@{}}
\toprule
Methods/Val.  & & 0.2$y$+0.3$x$. &0.4$y$+0.3$x$ \\ \midrule
\multirow{1}{*} Single-CE      
& Best  &57.84   &47.76    \\
& Last  &57.39   &41.39 \\
\hline
\multirow{1}{*}BNNs           
&Best &62.25  &51.20    \\
&Last &61.96  &50.68   \\
\hline 
\multirow{1}{*}SNGP   
& Best &59.92  &49.27 \\
& Last &59.22  &49.09\\
\hline
\multirow{1}{*}DE-CE           
& Best &\textbf{66.50}    &\textbf{54.90}    \\
& Last &\textbf{65.24}    &\textbf{53.91}\\
\bottomrule
\label{tab:append-uq}
\end{tabular}
\end{table}
We also explore more uncertainty-based models in addition to deep ensemble to fit the noisy data with dual noises, e.g., Bayesian neural networks (BNNs) with mean-field variational inference (MFVI) and Spectral-normalized Neural Gaussian Process (SNGP)~\cite{van2020uncertainty}. 
We do not consider the uncertainty-based models (e.g., Monte Carlo (MC) dropout~\cite{gal2016dropout}, DUQ~\cite{liu2020simple} and Prior Network~\cite{malinin2018predictive}) that can not explicitly model uncertainty at the training phase in our experiments because the lack of uncertainty can not alleviate over-fitting during the training time. Table~\ref{tab:append-uq} presents the classification accuracy of uncertainty-based models and deterministic DNNs on CIFAR-100 with dual noises. We can see that uncertainty-based models can better alleviate over-fitting than deterministic DNNs. Especially, deep ensemble used in this paper can achieve the best performance compared to BNNs and SNGP, so we opt to place our workflow on the well-evaluated deep ensemble to establish a strong learning approach under dual noises.
\begin{table}[!ht]
\centering
\setlength{\tabcolsep}{2.5mm}
\caption{The AUROC scores of detecting noisy samples with different levels of \emph{x-noise} and \emph{y-noise}.}
\begin{tabular}{@{}lcccc@{}}
\toprule
Noise rate & 0.3$x$ & 0.4$x$ & 0.3$y$ & 0.4$y$ \\ \midrule
\multirow{1}{*}       
AUROC  &89.10\%    &88.13\%     &95.21\%   & 94.02\%    \\
\bottomrule
\label{tab:auroc}
\end{tabular}
\end{table}
\section{Experiment details}
\label{appendix:exp}
\subsection{Preprocessing}
All images are normalized and augmented by random horizontal flipping. For CIFAR-100, we use the standard $32\times32$ random cropping after zero-padding with 4 pixels on each side. For TinyImageNet, we use randomly crop a patch of size $56\times56$ from each image. For WebVision, we first resize each image to make the size as 320. Then we use the standard data augmentation, randomly crop a patch of size $299\times299$ from each image, and apply horizontal random flipping.
\subsection{Optimizer and hyper-parameters} 
SGD with momentum (0.9) and weight decay $3\times10^{-4}$ is used in all experiments. For the setting of the thresholds $\epsilon_1$ and $\epsilon_2$, we recommend performing a grid search for $\epsilon_1 \in \{0.015, 0.020, 0.025, 0.030, 0.035\}$ and $\epsilon_2 \in \{1\%, 2\%, 3\%, 4\%, 5\%, 6\%\}$ to achieve the better performance. In the experiments, we set $\epsilon_1 = 0.020$ and $\epsilon_2 = 5\%$ for CIFAR-100 and TinyImageNet.
\subsection{Network}
Five networks with ResNet-18  are trained from scratch using PyTorch 1.9.0. for all experiments. Default PyTorch initialization is used on all layers. It is noteworthy that we need to use the small convolution with $3\times3$ kernel in the downsampling layer for CIFAR-100 and TinyImageNet.
\subsection{Warm-up}
The model warm-up can help better separate noisy data and clean data. We start training the model with high learning rates and standard cross-entropy loss in experiments. Specifically, our method uses the learning rate of 0.2 for the first 35 epochs for CIFAR-100 and TinyImageNet. For WebVision, we use the learning rate of 0.2 for the first 40 epochs.

\begin{figure}
  \centering
  \begin{subfigure}[1]{0.32\linewidth}
    \includegraphics[width=0.95\linewidth]{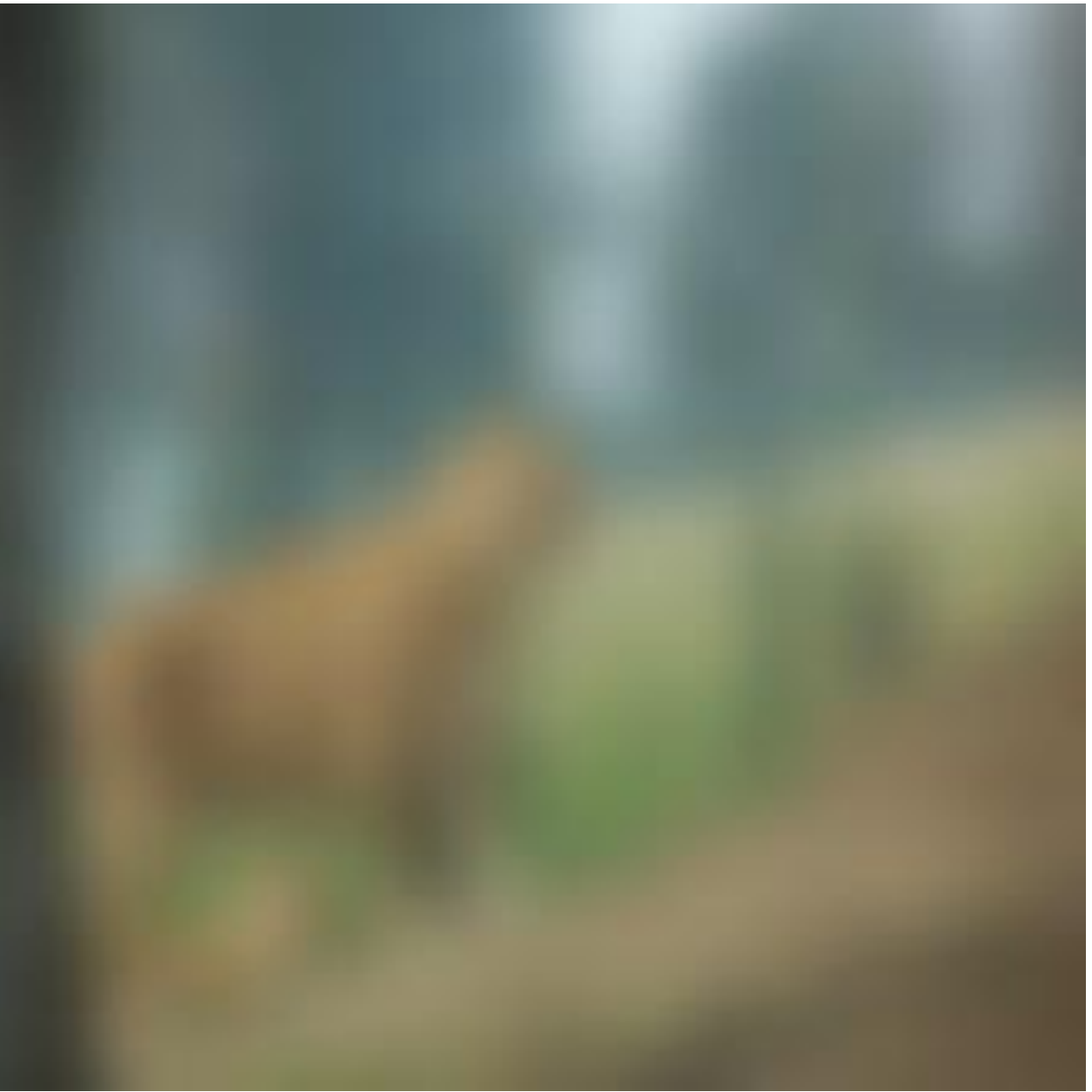}
    \caption{The corrupted image using the Gaussian blur.}
  \end{subfigure}%
  \hfill
  \begin{subfigure}[2]{0.32\linewidth}
    \includegraphics[width=0.95\linewidth]{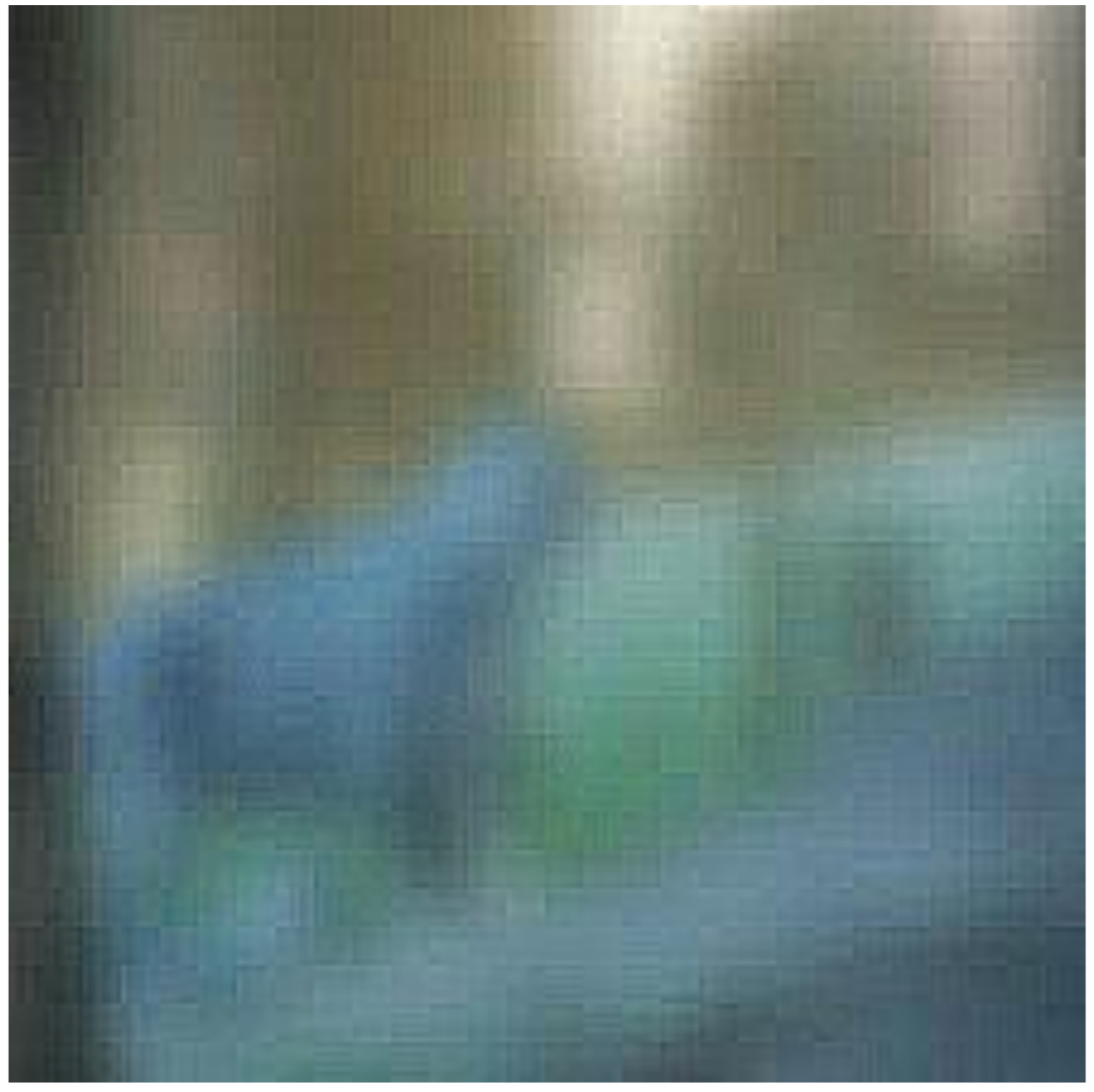}
    \caption{The restored image using the mean filter in \emph{OpenCV}.}
  \end{subfigure}
  \begin{subfigure}[3]{0.30\linewidth}
    \includegraphics[width=0.95\linewidth]{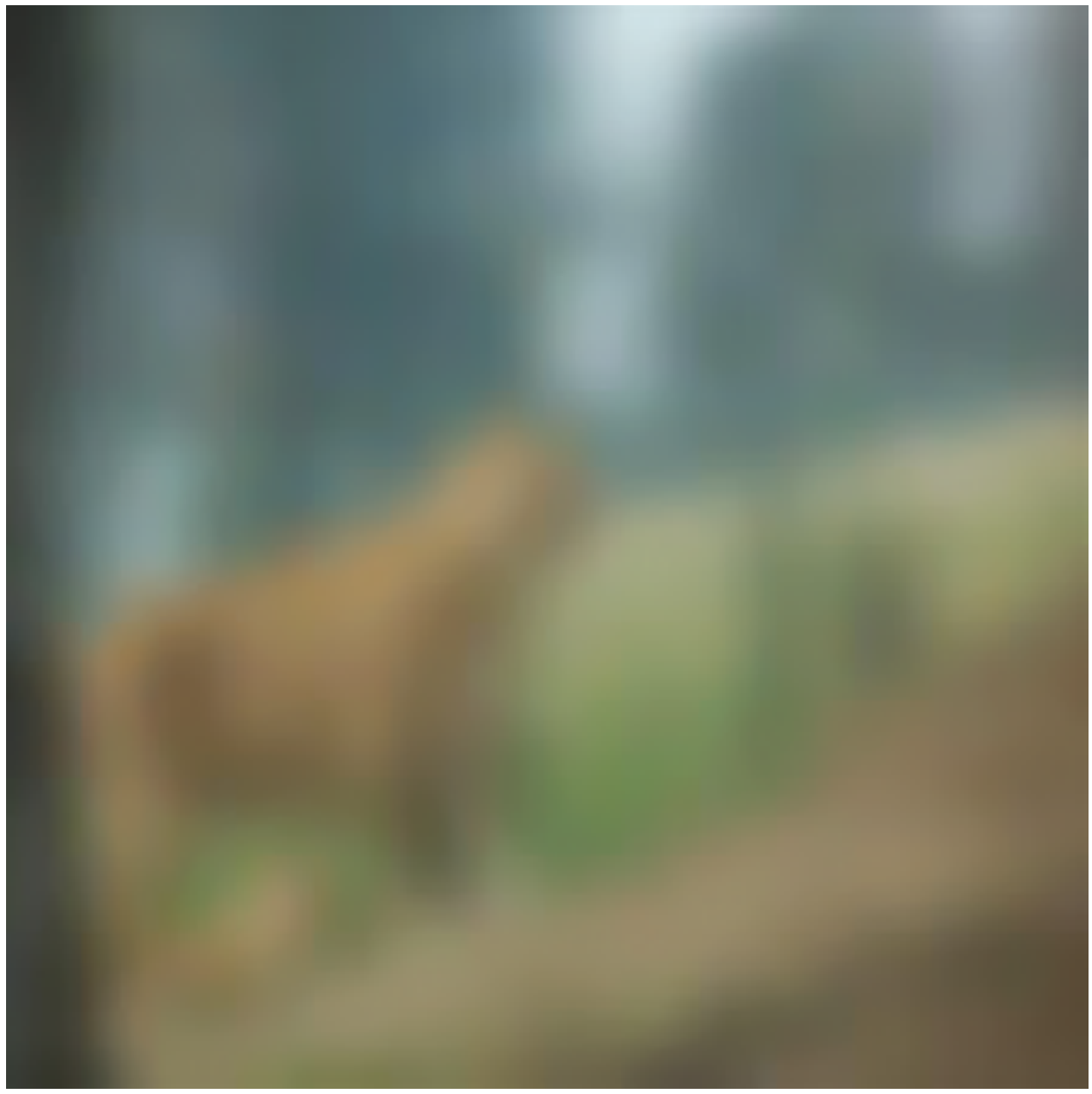}
    \caption{The restored image using the latest denoising technique based on NNs~\cite{carbajal2021single}.}
  \end{subfigure}
  \caption{Blurred and restored images using different image denoising methods.}
  \label{deblur}
\end{figure}

\subsection{Training schedule}
For CIFAR-100 and TinyImageNet, training for 250 epochs in total, and we reduce the initial learning rate (0.2) by a factor of 2.5 after 35, 80, 120, 150 and 180 epochs. For WebVision, we train the model for 130 epochs and reduce the initial learning rate (0.1) by a factor of 10 after 80 and 105 epochs.

\section{Performance under synthetic y-noise}
\label{appendix-y}
\begin{table}[ht]
\vspace*{-0.5cm}
\renewcommand\arraystretch{1.2}
\centering
\setlength{\tabcolsep}{1.8mm}
\caption{The comparison of validation accuracy on CIFAR-100 and TinyImageNet with \emph{y-noise}.}
\begin{tabular}{@{}lcccccc@{}}
\toprule
Alg./Noise rate &  & 0.0 & 0.1 & 0.2 & 0.4 & 0.6 \\ \midrule

& & \multicolumn{4}{c}{CIFAR-100 / TinyImageNet }   \\
\cline{3-7}

\multirow{1}{*} DE-CE       
& Best & 79.13/63.62  &75.30/60.03   & 71.19/55.82    & 60.65/47.89    &51.26/39.14    \\
&Last  & 77.01/61.28  &75.16/58.06   &70.84/53.02     & 59.16/41.00   & 42.74/32.36    \\
 \hline
\multirow{1}{*} DYR~\cite{arazo2019unsupervised}  
&Best & 78.64/\textbf{65.14}    &73.76/60.04   &68.47/56.33    &58.43/47.85     &46.02/37.19   \\
&Last & 78.02/63.97    &73.13/58.13   &67.31/54.28    &57.06/45.72     &44.91/35.86   \\
\hline
\multirow{1}{*} M-DYR~\cite{arazo2019unsupervised}   
& Best & 75.38/62.32   & 75.43/60.40  & 75.18/\textbf{59.59}    &69.43/54.59     & 59.48/42.06  \\
& Last & 74.91/61.04   & 75.12/59.11  & 74.69/\textbf{58.25}    &68.73/52.77     & 56.07/41.26\\
\hline
\multirow{1}{*} CORES$^2$~\cite{cheng2020learning}            
& Best &76.76/59.74    &71.79/57.15   &67.42/54.53     &55.18/46.95     &42.97/37.17  \\
& Last &76.22/59.14    &71.03/57.00   &66.62/53.26     &54.31/45.39     &41.89/36.02\\
\hline
\multirow{1}{*}Proposed-L 
& Best & 80.44/64.07   &\textbf{79.45}/60.25   &\textbf{75.76}/58.96    &69.44/59.02    & 58.76/43.00   \\
& Last & 79.03/63.21   &77.15/59.31   &74.89/58.87    &66.04/\textbf{55.98}    &57.02/41.54\\
\hline
\multirow{1}{*}Proposed-LM
&Best & \textbf{80.98}/64.58    &79.32/\textbf{60.73}   &75.20/58.65    &\textbf{70.77}/\textbf{56.21}    &\textbf{59.52}/\textbf{44.65}   \\
&Last & \textbf{79.71}/\textbf{64.15}    &\textbf{78.81}/\textbf{59.95}   &\textbf{75.15}/58.01     &\textbf{68.83}/55.84    &\textbf{56.26}/\textbf{44.10}\\ 
\bottomrule
\label{tab:noisy_y_c100}
\end{tabular}
\end{table}

We report the results of all methods only under label noise in order to directly compare the proposed method with previous works focusing on this setting. Table~\ref{tab:noisy_y_c100} presents the results on CIFAR-100 and TinyImageNet with different levels of \emph{y-noise}, which shows that the proposed method significantly outperforms competitive methods for noisy labels (i.e., DYR, M-DYR, and CORES$^2$). The results verify that our method is also suitable for scenarios that only involve \emph{y-noise}. In particular, the proposed methods (i.e., no matter Proposed-L or Proposed-LM) also exhibit superior performance on clean data(i.e., 0\% noise). The result illustrates the proposed sample filtering and learning strategy is robust and can not bias the learning of the model on overall clean datasets, which is not enjoyed by other methods.
\begin{figure*}[ht]
    \centering
    \includegraphics[width=0.97\linewidth]{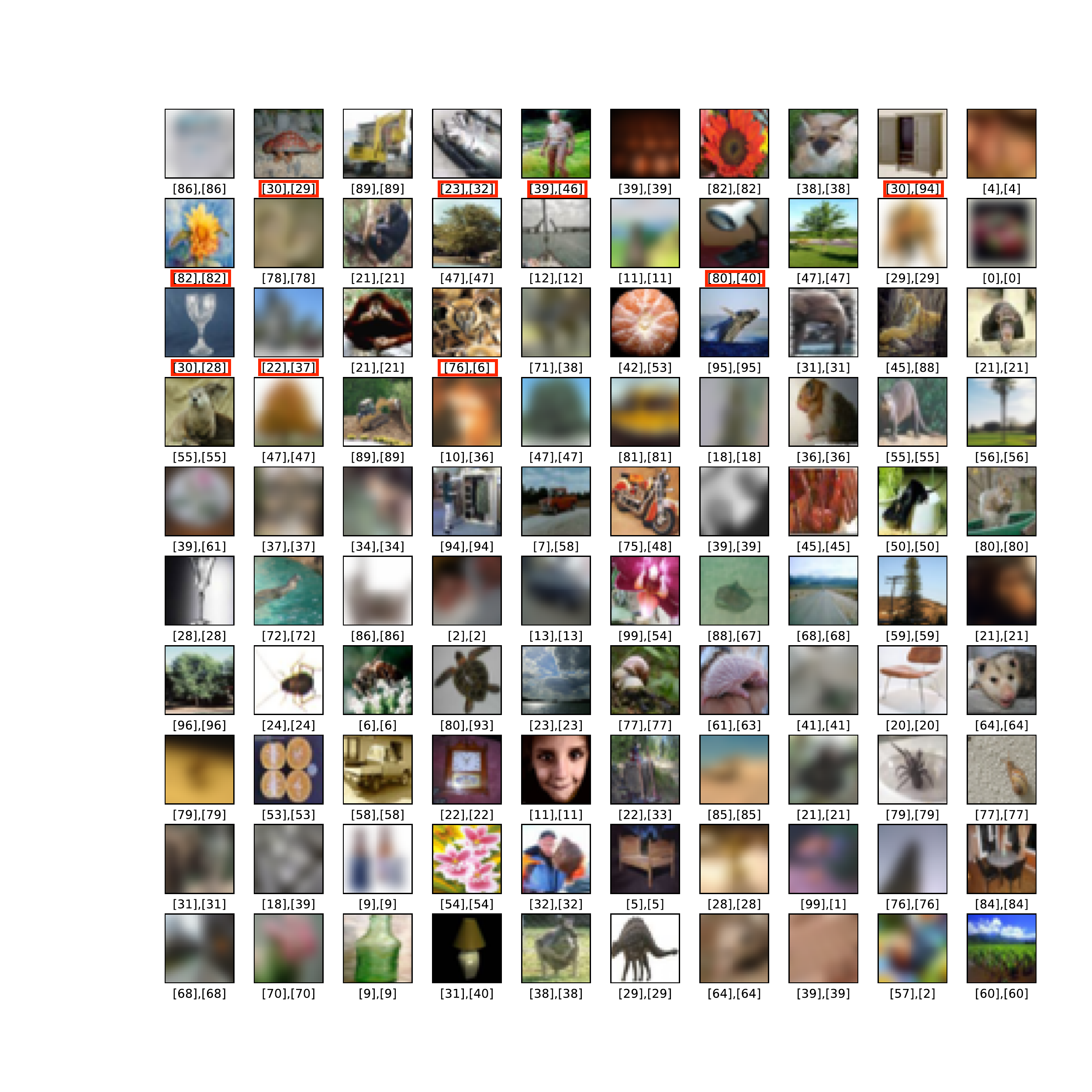}
    \caption{Some random images of CIFAR-100 with dual noises. The two numbers below each image represent the actual label id and the correct label id respectively. If two ids are not identical, it indicates an image with label noise.}
    \label{fig:exp_data}
\end{figure*}

\begin{figure*}[ht]
    \centering
    \includegraphics[width=0.97\linewidth]{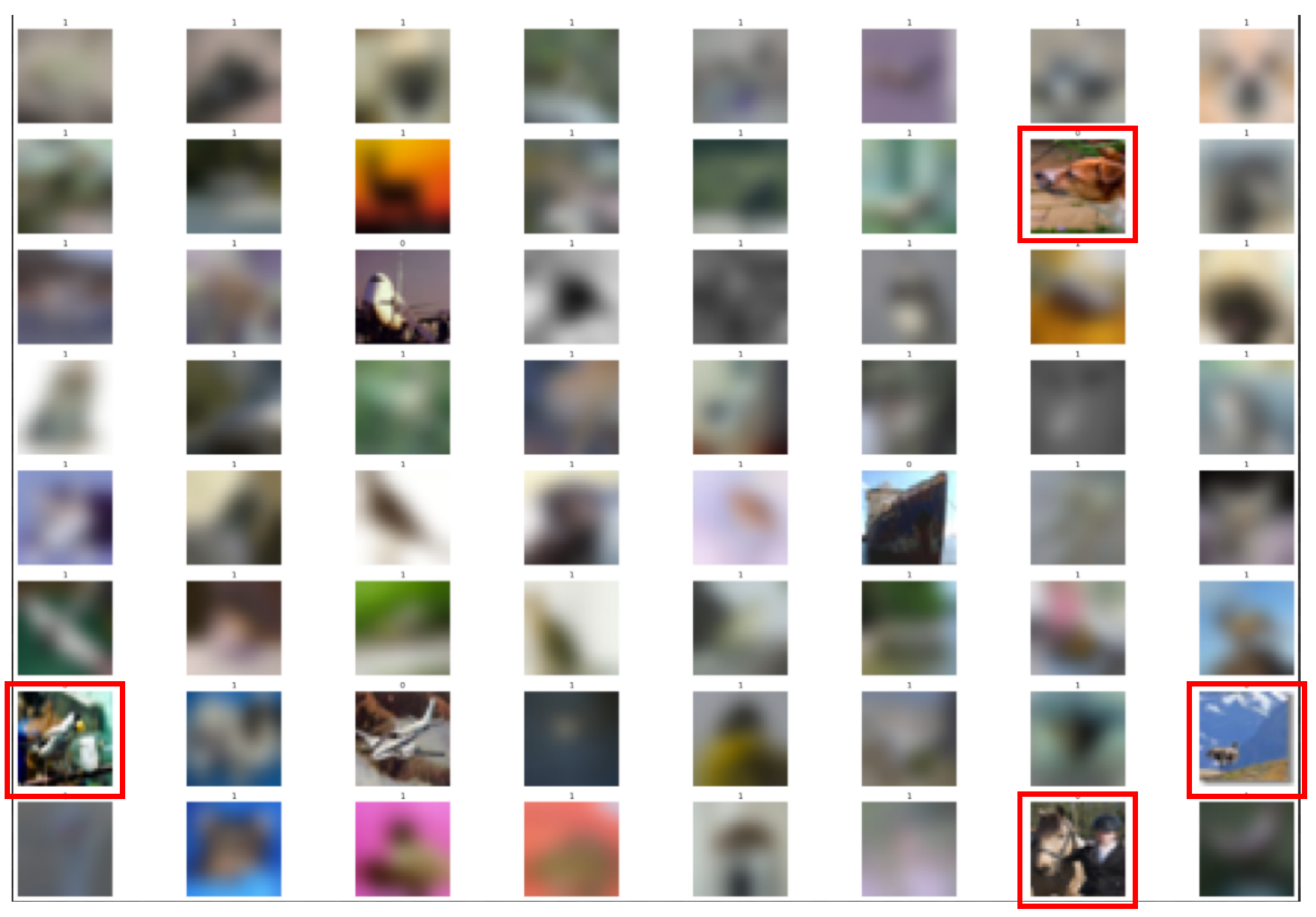}
    \caption{There are some low \textbf{M-Cons} corresponding images. Images with red boxes represent hard or dirty samples (e.g., some images contain multiple objects or some images contain background noise).}
    \label{fig:bias_data}
\end{figure*}
\end{document}